\documentclass[10pt,twocolumn,letterpaper]{article}
\usepackage[pagenumbers]{cvpr}
\usepackage[T1]{fontenc}
\usepackage{cuted,placeins}
\definecolor{cvprblue}{rgb}{0.21,0.49,0.74}
\usepackage[breaklinks,colorlinks,allcolors=cvprblue]{hyperref}
\newcommand{\irYes}{\ensuremath{\checkmark}}
\newcommand{\irNo}{\ensuremath{\times}}
\newcommand{\irPartial}{\ensuremath{\circ}}
\title{RoboInF: A Benchmark for Instruction Fidelity in Robotic Manipulation}
\author{Jiawei Guo$^{1*}$ \quad Yu Huang$^{1*}$ \quad Dijiang Dong$^{1*}$ \quad
Yue Liu$^{2}$ \quad Peiran Ren$^{2\dagger}$ \quad Wei Shen$^{1\dagger}$\\[3pt]
$^1$Shanghai Jiao Tong University \qquad $^2$Tongji University}
\hypersetup{pdftitle={RoboInF: A Benchmark for Instruction Fidelity in Robotic Manipulation},
pdfauthor={Jiawei Guo, Yu Huang, Dijiang Dong, Yue Liu, Peiran Ren, Wei Shen}}
\begin{document}
\maketitle
\begin{strip}
\centering
\includegraphics[width=\linewidth]{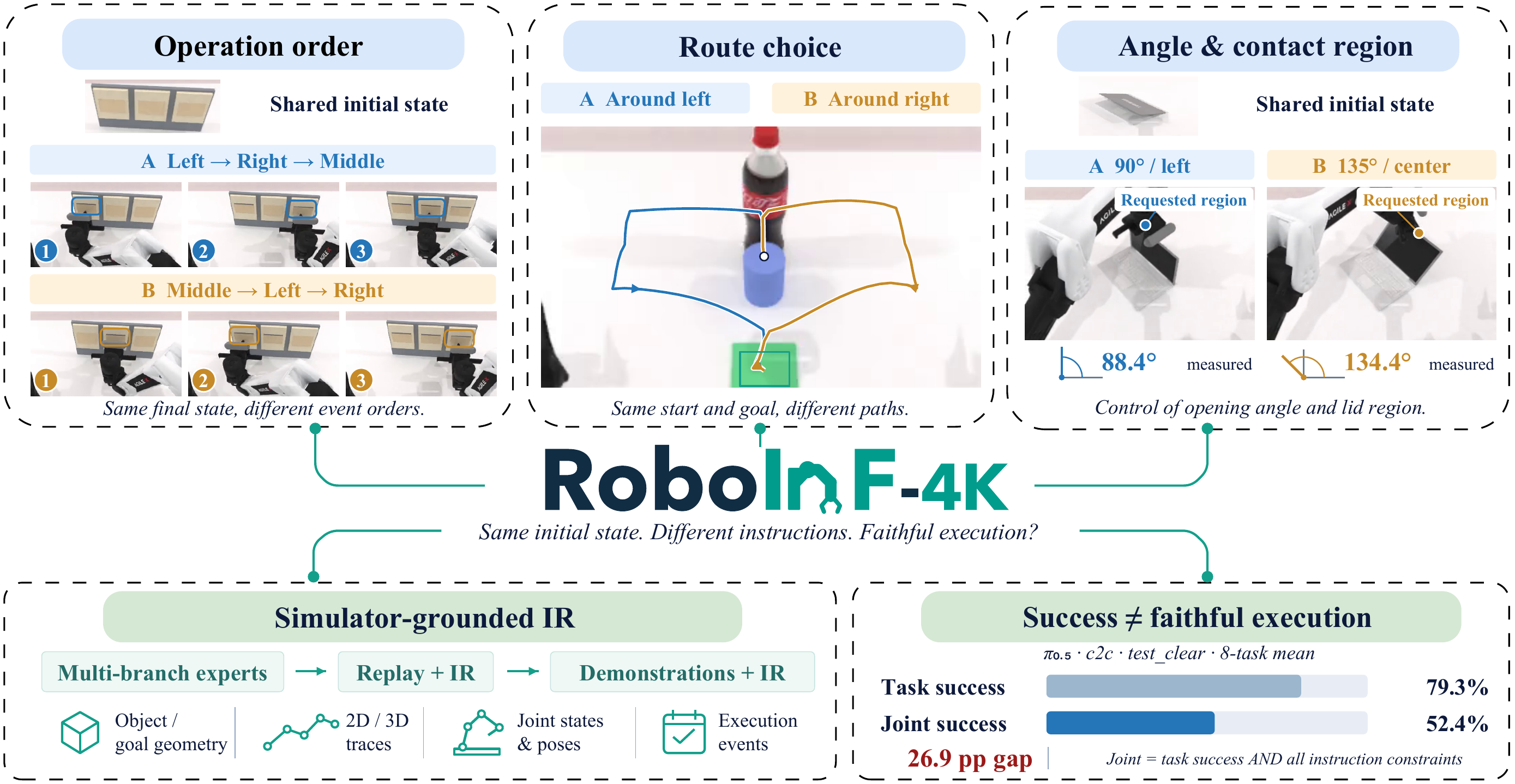}
\captionsetup{hypcap=false}
\captionof{figure}{\textbf{RoboInF evaluates instruction fidelity beyond task completion.}
From the same initial state, instructions specify switch order, route, opening angle, or lid contact region. Paths overlay the initial scene, and lid markers denote target contact centers.
RoboInF-4K pairs expert demonstrations with simulator annotations. Joint success requires completing the task and satisfying every instruction constraint.}
\label{fig:teaser}
\captionsetup{hypcap=true}
\end{strip}

\begingroup
\renewcommand{\thefootnote}{\fnsymbol{footnote}}
\footnotetext[1]{Equal Contribution.\quad\textsuperscript{\dag}Corresponding authors.}
\endgroup
\begin{abstract}
Precise language control requires embodied models to follow instructions governing both task outcomes and execution.
Although recent methods train policies to follow increasingly detailed commands, existing benchmarks often assess instruction following alongside perception, planning, and generalization to unfamiliar scenes.
Higher task success therefore offers limited evidence of closer adherence to instructions.
We introduce \textbf{RoboInF}, a benchmark measuring how reliably manipulation policies adapt their behavior to changes in an instruction.
For each of eight tasks, six to twelve expert-validated instruction branches vary the requested goal, procedure, or manipulation detail while holding the initial scene and robot state fixed.
Task-specific checkers verify the terminal goal and use recorded trajectories, contacts, and event sequences to assess compliance throughout execution.
The accompanying \textbf{RoboInF-4K} dataset pairs 4,242 expert demonstrations with simulator-derived annotations of motion and contact.
We evaluate leading manipulation policies ($\pi_{0.5}$ and OpenWAM) and inference-only agents (GPT-6 Astra and Qwen3.8-Flash).
All four models exhibit a gap between task completion and faithful execution, indicating substantial room for fine-grained instruction following.
\end{abstract}

\section{Introduction}
\label{sec:intro}

Embodied manipulation benchmarks such as LIBERO~\cite{libero}, RoboTwin~\cite{robotwin,robotwin2}, and RoboCasa~\cite{robocasa} have helped establish a common evaluation ecosystem for language-conditioned robot manipulation. They provide diverse tasks and environments for measuring capabilities such as task execution, generalization, and object interaction, and have become important testbeds for evaluating vision--language--action (VLA) models~\cite{rt2,openvla} and emerging world-action models~\cite{openwam_pretraining,openwam_composable}. Recent benchmark audits also caution that high aggregate scores can reflect shortcuts rather than the intended capabilities~\cite{manipulation_benchmark_audit}. Yet a successful task outcome does not necessarily mean that a policy has followed the instruction faithfully. A policy may reach the required final configuration while performing operations in the wrong order, taking an unintended route, contacting a region other than the one requested, or stopping at an imprecise target angle. Such executions may satisfy a coarse task-success criterion while violating the behavior explicitly requested by the instruction. This raises a complementary question for embodied evaluation: beyond whether a robot completes a task, can we measure whether it completes the task in the way it was instructed to?

Recent work has increasingly explored how to make robot manipulation policies more controllable through fine-grained language instructions, allowing users to specify not only what the robot should achieve, but also how it should execute the task~\cite{steerable_policies,finevla,resteer}. However, evaluating this capability remains challenging, as existing benchmarks often entangle instruction following with other capabilities, including object generalization, visual recognition, and multi-step planning. For example, InstructVLA~\cite{instructvla} evaluates instruction tuning on SimplerEnv-Instruct, where performance is also affected by unfamiliar objects and visual distractors, while VLABench~\cite{vlabench} includes tasks requiring implicit goal inference and multi-step planning. Consequently, improvements in task success may arise from better perception or planning, rather than a stronger ability to execute instructions faithfully. This makes it difficult to isolate whether a policy actually responds to the specific behavioral constraints expressed in language. We therefore need a dedicated benchmark that isolates instruction fidelity by testing whether, under otherwise identical conditions, a robot adjusts its actions as a revised instruction requires.

To make such an evaluation possible, we argue that a benchmark for instruction fidelity should meet two requirements. First, it should compare different instructions from the same initial scene and robot state. If the scene or initial conditions change together with the instruction, differences in execution may reflect changes in perception, object configuration, or task difficulty rather than the policy's response to language. We therefore require each initial state to admit multiple feasible instruction branches that vary the requested goal, procedure, or manipulation details. Second, evaluation should check the parts of execution that the instruction actually constrains, rather than relying only on final-state success. This is essential when different procedures lead to the same final configuration, or when a task is considered complete despite imprecise execution. Figure~\ref{fig:teaser} illustrates both cases: different switch orders yield the same final state, while a laptop can reach the requested angle even if the robot contacts the wrong part of the lid. Together, these requirements let us assess whether a policy's behavior changes appropriately with the instruction and whether the resulting execution satisfies its constraints.

Following these principles, we introduce \textbf{RoboInF}, a benchmark for evaluating instruction fidelity in language-conditioned manipulation, built on RoboTwin~\cite{robotwin,robotwin2}. RoboInF contains eight base tasks, each with six to twelve instruction branches from shared initial states. To generate demonstrations that satisfy these branch-specific requirements, we extend RoboTwin's vision--language model (VLM)-based expert policy generator to synthesize a multi-branch expert policy for each task. Each policy program takes a branch specification as input and executes the corresponding goal, procedure, and fine-grained requirements from the same initial state. We refine these programs iteratively using execution videos and structured simulator logs, together with task-specific compliance checks. During data collection, a structured intermediate representation (IR) collector extracts task and interaction annotations directly from simulator states and execution records, avoiding video-based estimation and manual label verification. The resulting \textbf{RoboInF-4K} dataset contains post-training demonstrations with detailed IR annotations, supporting evaluation of task-only policies and policies trained with IR-derived commands using the same branches and checks.

With the benchmark and dataset in place, we study instruction fidelity across different policy settings. We evaluate VLA policies in terms of both task completion and compliance with instruction-specific requirements, and examine post-training and an IR-supported hierarchical control configuration. We also evaluate inference-only agents separately, accounting for their different action interfaces. By distinguishing final task outcomes from the goals, procedures, and manipulation details specified by an instruction, RoboInF provides a common testbed for studying how reliably embodied policies can be controlled through language.

\section{Related Work}
\label{sec:related_work}

\noindent\textbf{Manipulation Benchmarks and Datasets.}
SAPIEN, RLBench, and ManiSkill2 provide simulation environments for robotic manipulation~\cite{sapien,rlbench,maniskill2}. CALVIN~\cite{calvin} evaluates long-horizon language control, and LIBERO~\cite{libero} tests knowledge transfer across tasks. RoboDojo~\cite{robodojo} evaluates instruction following alongside precision and memory in simulation and on real robots. Real-world datasets broaden scene and task coverage~\cite{droid,bridgedata_v2}, while Open X-Embodiment aggregates demonstrations across robot embodiments~\cite{open_x_embodiment}. Trajectory adaptation~\cite{mimicgen}, household simulation~\cite{robocasa}, and generative digital twins~\cite{robotwin,robotwin2} have expanded the range of available demonstrations. GenSim and GenSim2 generate simulation tasks and demonstrations through language-model-generated programs~\cite{gensim,gensim2}, while RoboGen automates task, scene, and supervision generation~\cite{robogen}. Complementing these efforts to scale robotic experience, we build on RoboTwin's expert generator to validate multiple feasible instructions for each initial scene.

\noindent\textbf{Evaluating Instruction Fidelity.}
GENMANIP~\cite{genmanip} tests generalization across scenes, VLABench~\cite{vlabench} evaluates long-horizon reasoning, and InstructVLA~\cite{instructvla} trains policies to generalize to new instructions. Even when policies complete a task, their actions can remain insensitive to changes in the instruction~\cite{icbench,encoded_not_control}. Perturbation-based evaluations further expose sensitivity to scene configurations and weak dependence on language instructions~\cite{libero_plus,libero_pro}. LIBERO-Para evaluates robustness to meaning-preserving paraphrases~\cite{libero_para}. LIBERO-CF~\cite{libero_cf} tests alternative instructions in shared layouts, while InstructMove~\cite{instructmove} uses semantic distractors to make target selection depend on language. Other evaluations probe implicit references~\cite{roboirgbench}, exclusions~\cite{negabench}, sensitivity at individual steps~\cite{language_matter}, and responses to mid-execution instruction changes~\cite{resteer}. SIMPLER-IN~\cite{simpler_in} measures whether successful executions satisfy path, grasp, or placement requirements. RoboInF pairs expert-validated branches with execution checks for arm use, contact, and operation order, including cases where different procedures yield the same final state.

\noindent\textbf{Structured Representations for Manipulation.}
Prior work represents geometric objectives with keypoints~\cite{kpam}, spatial affordances~\cite{robopoint,moka}, and value maps~\cite{voxposer}, or encodes spatial structure in policy inputs and actions~\cite{spatialvla}. Programs~\cite{code_as_policies} and relational constraints~\cite{rekep} guide execution, while visual monitors check constraints during manipulation~\cite{code_as_monitor}. VIMA expresses manipulation tasks through interleaved textual and
visual prompts~\cite{vima}. Goal sketches specify desired configurations~\cite{rt_sketch}, whereas trajectory sketches and predicted end-effector paths provide motion guidance~\cite{rt_trajectory,hamster}. Other policies use language motions~\cite{rth}, reasoning traces~\cite{ecot}, or commands at multiple levels of detail~\cite{steerable_policies}. RoboInter provides intermediate annotations for manipulation videos~\cite{robointer}, and FineVLA constructs human-verified, fine-grained instruction supervision~\cite{finevla}. Complementing these representations and annotation pipelines, RoboInF records geometry, motion, and contact directly from the simulator and evaluates the resulting systems against explicit instruction constraints.

\section{Method}
\label{sec:method}

We first define instruction fidelity through the requirements governing an execution and the controlled comparisons used to assess it. We then describe the data engine that constructs multi-branch experts, collects structured intermediate annotations, and assembles expert-validated demonstrations for RoboInF. Detailed task designs, canonical instructions, and task-specific evaluation criteria are provided in Appendix~\ref{sec:task_design_details}.

\subsection{Definition of Instruction Fidelity}
\label{sec:fidelity_definition}

\begin{figure*}[t]
    \centering
    \includegraphics[width=\linewidth]{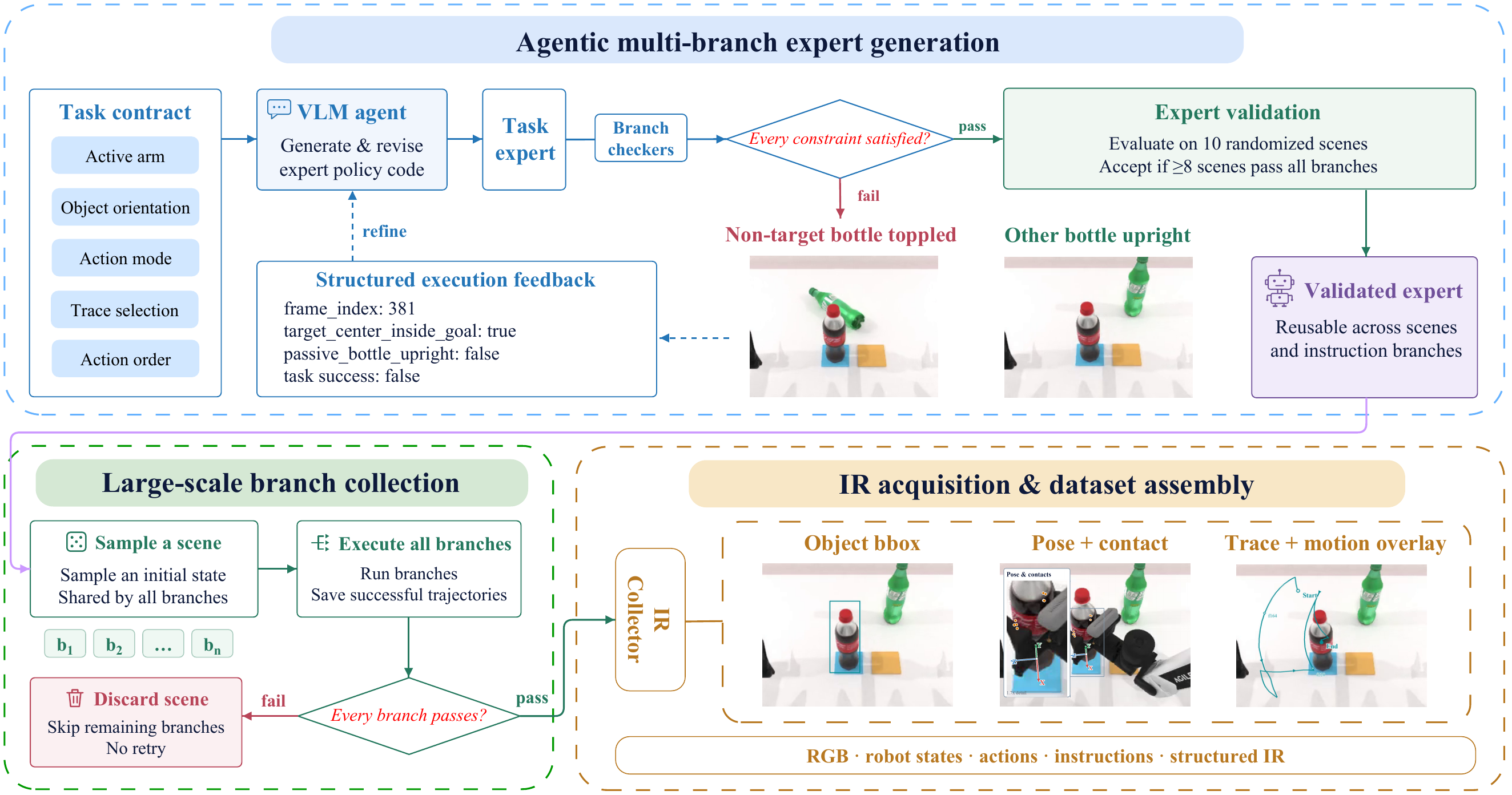}
    \caption{\textbf{Data collection engine.}
A VLM iteratively refines a multi-branch expert using execution videos and structured feedback.
Scene-level validation retains only complete, successful branch sets, with $n$ denoting the task-specific number of branches (6--12).
Replay synchronizes demonstrations with simulator-ground-truth IR.
The bottle examples contrast a toppled non-target bottle with a successful placement; the IR visualizations show object bounds, pose and contacts, and recorded object motion.}
    \label{fig:data_collection_engine}
\end{figure*}

\paragraph{Problem formulation.}
Let $g$ denote a base manipulation task, $s_0$ the complete initial simulator state, and $c=(c_1,\ldots,c_m)$ a specification of instruction conditions. The corresponding natural-language instruction is $\ell(g,c)$. At step $t$, a policy $\pi$ receives visual and proprioceptive observations $o_t$ and selects an action $a_t$ conditioned on the instruction and observation history. An execution with at most $H$ action steps is
\begin{equation}
    \tau=(s_0,a_0,s_1,\ldots,a_{L-1},s_L),\qquad L\leq H.
\end{equation}
The task instruction remains fixed throughout the episode. We define \emph{instruction fidelity} as satisfaction of the conditions specified by that instruction, including requirements on the final configuration and the execution leading to it. For example, placing a bottle at its destination does not establish compliance with an instruction to use the left arm and pass around the left side of a divider.

\begin{figure*}[t]
    \centering
    \includegraphics[width=\textwidth]{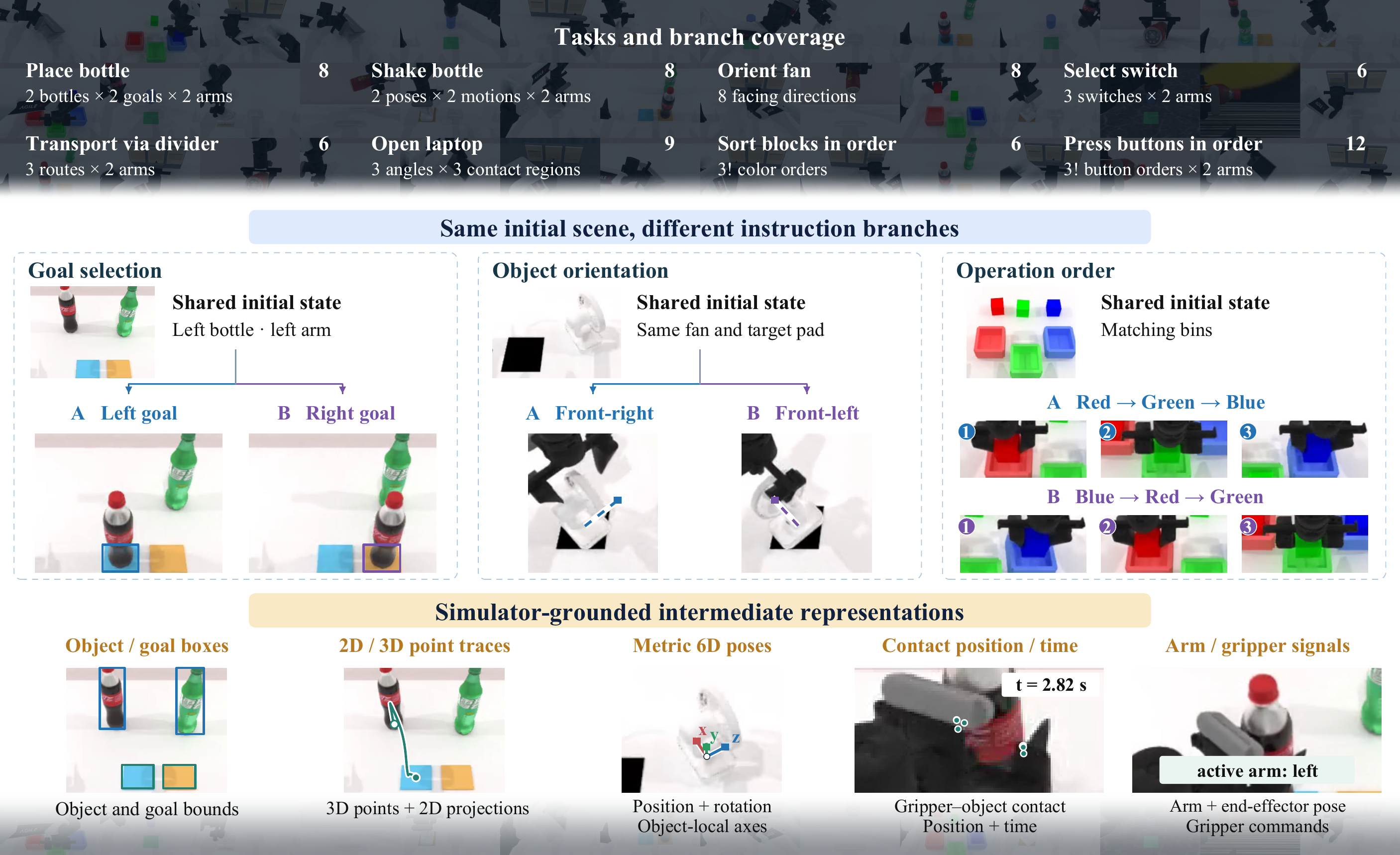}
    \caption{\textbf{RoboInF-4K dataset overview.}
    Top: eight tasks with six to twelve branches per scene.
    Middle: demonstrations follow different instructions from a shared initial state, varying the target goal, object orientation, or execution order; dashed arrows indicate requested table-frame directions.
    Bottom: simulator-grounded intermediate representations (IR), including visible boxes, measured 3D traces with 2D projections,
    6D poses, actual contact points/times, and arm information with gripper commands.
    Requested conditions and observed outcomes are distinguished.}
    \label{fig:dataset_overview}
\end{figure*}

\paragraph{Factorized instruction constraints.}
We organize instruction conditions around the three aspects introduced above: \emph{goal selection}, \emph{procedural adherence}, and \emph{fine-grained specifications}. Goal selection identifies the object, destination, or articulated component to operate. Procedural conditions specify arm assignment, routes, motion patterns, repetition counts, or operation orders. Fine-grained specifications constrain quantities such as object orientation, holding pose, joint angle, and contact region. These aspects can coexist within one instruction. Figure~\ref{fig:dataset_overview} shows which instruction variables change across branches and how many branches each task admits. These variables specify the requested behavior.

\paragraph{Task success and faithful execution.}
For each task and branch, let $G_{g,c}$ be a binary terminal-goal predicate. We define task success by
\begin{equation}
    T(\tau;g,c)=G_{g,c}(s_L)\in\{0,1\}.
    \label{eq:terminal_success}
\end{equation}
The terminal goal depends on the branch, such as the selected object or destination; task-specific predicates are listed in Appendix~\ref{sec:task_design_details}.

Let $\mathcal K(g,c)$ index the applicable instruction constraints, and let $C_k(\tau;g,c)\in\{0,1\}$ indicate whether execution satisfies constraint $k$. Each $C_k$ evaluates a requirement expressed in the instruction using terminal geometry, contact or activation events, or trajectory statistics. We define
\begin{align}
    F(\tau;g,c) &= \bigwedge_{k\in\mathcal K(g,c)} C_k(\tau;g,c), \label{eq:fidelity}\\
    J(\tau;g,c) &= T(\tau;g,c)\land F(\tau;g,c). \label{eq:joint_success}
\end{align}
Here, $F$ denotes constraint satisfaction, while $J$ requires both task completion and faithful execution. An execution that reaches the destination using the wrong arm can satisfy $T$ while failing $F$ and $J$.

\paragraph{Feasible same-state interventions.}
Let \(\mathcal B_g(s_0)\) denote the finite set of instruction branches associated with task \(g\) and initial state \(s_0\). Each branch specifies a distinct combination of instruction conditions and is evaluated in a separate rollout initialized from the same \(s_0\). Across branches, the initial scene, object states, and robot state remain fixed, while the requested goals, procedures, or fine-grained requirements vary.

Every retained scene has an expert execution witnessing feasibility for each branch:
\begin{equation}
    \forall c\in\mathcal B_g(s_0),\quad
    J(\tau^{E}_{s_0,c};g,c)=1,
    \label{eq:expert_feasibility}
\end{equation}
where $\tau^{E}_{s_0,c}$ starts from $s_0$ and obeys the execution budget. These branches enable instruction-conditioned comparisons from identical initial states, including procedures that share the same terminal configuration.

\subsection{Data Collection Engine}
\label{sec:data_engine}

Figure~\ref{fig:data_collection_engine} summarizes the data collection engine, from expert program refinement to validated demonstrations with structured IR annotations.

\paragraph{Task contracts and branch enumeration.}
Each task contract defines the independent instruction variables, their admissible combinations, and the predicates used to check terminal outcomes and execution constraints. Scene construction is separated from branch execution so that changing an instruction does not alter the initial physical state. From a shared branch specification, we derive a canonical instruction, a branch identifier, and the requested-condition fields of the structured IR. Execution-dependent annotations are populated during collection. Enumerating valid combinations yields six to twelve branches per task, with fixed or expert-resolved attributes distinguished from independently varied conditions. We specify the task contracts and checkers before refinement. The VLM iteratively updates the expert policy against these fixed predicates, thresholds, and acceptance criteria.

\paragraph{VLM-based agentic expert policy generation.}
Language-model-generated programs provide an executable interface to robot control~\cite{code_as_policies,gensim2}, while execution feedback can support closed-loop task planning~\cite{inner_monologue}. Here, feedback is used offline to revise a reusable expert program before dataset collection. We extend RoboTwin's VLM-based expert policy generator~\cite{robotwin,robotwin2} to synthesize one \emph{multi-branch expert policy} per task. The VLM receives the task description, instruction conditions, scene information, and robot manipulation interfaces, and produces an executable policy program conditioned on the branch specification. Executing a candidate in simulation returns two complementary feedback channels: execution videos and structured JSON logs. Videos expose the observed motion and interactions, while structured logs report states, events, checker outcomes, and associated errors. The VLM uses this feedback to diagnose failures and revise the program's action stages, contact choices, motion paths, or control parameters.

Candidate programs are assessed on ten development scenes, each evaluated over its complete branch set. A scene passes only if every branch satisfies $J$. Refinement runs for at most ten rounds. A program is accepted once at least eight of the ten scenes pass, corresponding to a scene success rate of at least $80\%$. We retain validated programs with their versions and refinement records for subsequent demonstration collection.

\begin{table*}[t]
    \centering
    \small
    \setlength{\tabcolsep}{5pt}
    \caption{Dataset statistics and expert-policy generation. Candidate scenes count attempted initial states; eligible scenes pass every instruction branch. Expert success is the scene eligible rate. Split sizes count episodes, although partitioning is by scene. Iterations exclude the initial program generation.}
    \label{tab:collection_statistics}
    \begin{tabular}{@{}lrrrrrrrc@{}}
        \toprule
        & \multicolumn{2}{c}{Scenes} & & & \multicolumn{3}{c}{Episodes} & \\
        \cmidrule(lr){2-3}\cmidrule(lr){6-8}
        Task & Candidate & Eligible & Branches / scene & Eligible rate (\%) & Train & Val & Test & Iterations \\
        \midrule
        Place bottle       & 287 & 63 & 8  & 21.9 & 296 & 104 & 104 & 10 \\
        Shake bottle  & 75  & 63 & 8  & 84.0 & 296 & 104 & 104 & 8  \\
        Orient fan  & 72  & 63 & 8  & 87.5 & 296 & 104 & 104 & 5  \\
        Select switch        & 109 & 84 & 6  & 77.1 & 300 & 102 & 102 & 2  \\
        Transport via divider      & 210 & 94 & 6  & 44.8 & 336 & 114 & 114 & 10  \\
        Open laptop        & 87  & 64 & 9  & 73.6 & 342 & 117 & 117 & 2  \\
        Sort blocks in order   & 100 & 89 & 6  & 89.0 & 318 & 108 & 108 & 5  \\
        Press buttons in order & 66  & 46 & 12 & 69.7 & 336 & 108 & 108 & 10  \\
        \midrule
        Total & 1006 & 566 & -- & 56.3 & 2520 & 861 & 861 & -- \\
        \bottomrule
    \end{tabular}
    \par\smallskip
    {\footnotesize The total eligible rate pools scene counts across tasks; branch counts and refinement iterations are task-specific.}
\end{table*}

\begin{table*}[t]
    \centering
    \small
    \setlength{\tabcolsep}{5pt}
    \caption{Supported annotation types across IR acquisition pipelines. \irYes: supported; \irPartial: partially supported; \irNo: unsupported by the specified pipeline. Manual labeling refers to per-episode annotation. Video methods infer annotations from observations; ours accesses simulator states and execution records.}
    \label{tab:ir_capabilities}
    \begin{tabular}{@{}lcccccc@{}}
        \toprule
        Pipeline & \shortstack{2D target\\localization} & \shortstack{2D point\\trajectories} & \shortstack{3D point\\trajectories} & \shortstack{Contact position\\and time} & \shortstack{Metric\\6D poses} & \shortstack{No manual\\labeling} \\
        \midrule
        GPT-6 Astra video prompting$^{d}$ & \irYes & \irPartial & \irNo & \irYes & \irNo & \irYes \\
        Grounded-SAM-2 & \irYes & \irNo & \irNo & \irNo & \irNo & \irYes \\
        Grounded-SAM-2 + CoTracker3 & \irYes & \irYes & \irNo & \irNo & \irNo & \irPartial$^{a}$ \\
        RoboInter-Tools & \irYes & \irNo & \irNo & \irPartial$^{b}$ & \irNo & \irNo \\
        Ours & \irYes & \irYes$^{c}$ & \irYes & \irYes & \irYes & \irYes \\
        \bottomrule
    \end{tabular}
    \par\smallskip
    \begin{minipage}{\textwidth}
        \footnotesize
        $^{a}$Tracking is automatic after query-point initialization, which may require manual selection. CoTracker3~\cite{cotracker3} outputs 2D image coordinates; 3D reconstruction requires additional depth or multiview information. Mask tracking alone~\cite{grounded_sam2} does not provide fixed-keypoint trajectories.
        $^{b}$RoboInter-Tools~\cite{robointer} includes human contact-frame annotation, which partially covers contact position and time. The RoboInter row refers only to its public Tools workflow.
        $^{c}$Our 2D trajectories are obtained by projecting recorded 3D keypoints using camera calibration.
        $^{d}$Point trajectories comprise sampled 2D landmarks; contact positions are image-plane estimates.
    \end{minipage}
\end{table*}

\paragraph{IR acquisition module.}
During execution and replay, a dedicated collector extracts ground-truth annotations of object geometry, robot motion, contacts, and task events directly from simulator states and execution records. The IR includes object and destination bounding boxes, object and end-effector poses, contacts, keypoint traces, arm identities, and task events. It stores branch targets alongside measured poses, contacts, and trajectories. Coordinate frames, units, frame indices, simulation steps, and timestamps support alignment with images, robot states, and actions. Validity masks identify missing or inapplicable annotations.

We store the annotations as structured metadata and trajectory arrays indexed by scene, branch, and episode, and synchronize them with observations and actions. Direct simulator access provides these geometric and interaction fields without video-based estimation or manual label verification, as shown in Table~\ref{tab:ir_capabilities}.

\paragraph{Large-scale collection and quality assurance.}
Validated experts are applied to sampled scenes and their instruction branches through a scan--replay pipeline. The scan stage records expert trajectories that pass the task and instruction checks. We apply a scene-level fail-fast rule: once any branch fails, the entire scene is discarded, its remaining branches are skipped, and that scene is not retried. Scenes that pass the full branch set are reconstructed for trajectory replay and synchronized recording of RGB observations, robot states, actions, camera information, and IR annotations. Initial states are verified across branches and before replay, and execution checks are repeated after replay. Scenes failing replay checks are excluded.

Final quality assurance checks behavioral compliance, branch completeness, initial-state consistency, agreement between instructions and conditions, and temporal alignment of videos, actions, and IR. Complete validated scene groups form RoboInF-4K, with all sibling branches assigned to the same dataset split.

\section{Experiments}
\label{sec:experiments}

We evaluate instruction fidelity in learned manipulation policies and inference-only agents on our RoboInF benchmark. Additional evaluation results are provided in Appendix~\ref{sec:zero_shot_ood}.

\subsection{Experimental Setup}
\label{sec:experimental_setup}

\begin{figure*}[t]
\centering
\setlength{\tabcolsep}{1pt}
\newlength{\appearancePanelWidth}
\setlength{\appearancePanelWidth}{\dimexpr(\textwidth-14\tabcolsep)/8\relax}
\begin{tabular}{@{}*{8}{c}@{}}
\scriptsize Place bottle & \scriptsize Shake bottle & \scriptsize Orient fan & \scriptsize Select switch & \shortstack{\scriptsize Transport via\\\scriptsize divider} & \scriptsize Open laptop & \shortstack{\scriptsize Sort blocks\\\scriptsize in order} & \shortstack{\scriptsize Press buttons\\\scriptsize in order} \\
\includegraphics[width=\appearancePanelWidth]{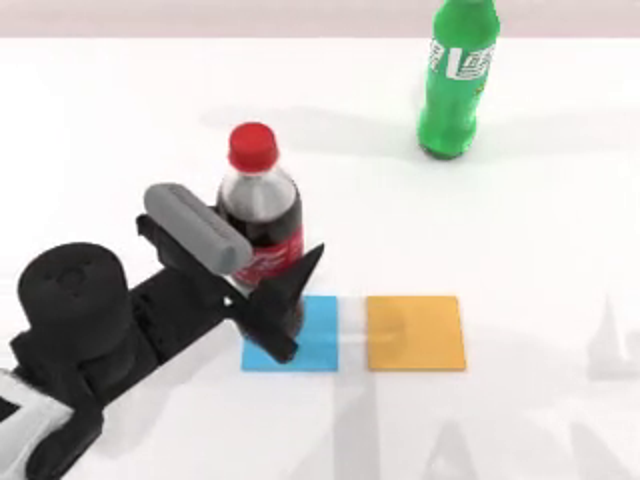} & \includegraphics[width=\appearancePanelWidth]{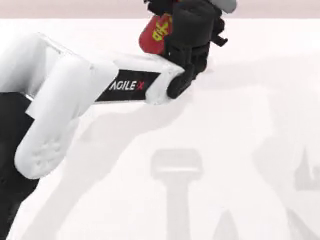} & \includegraphics[width=\appearancePanelWidth]{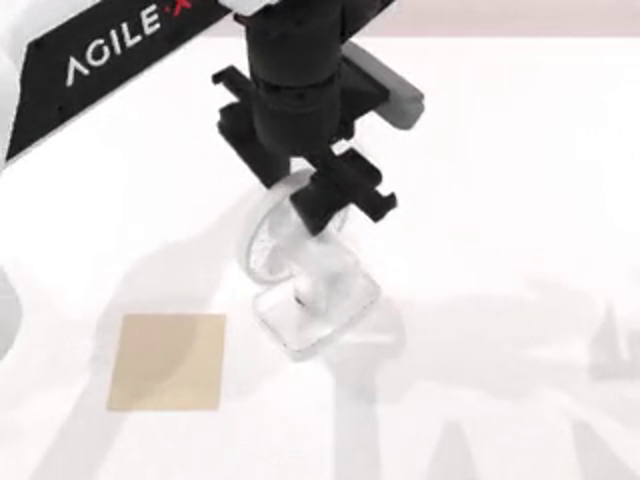} & \includegraphics[width=\appearancePanelWidth]{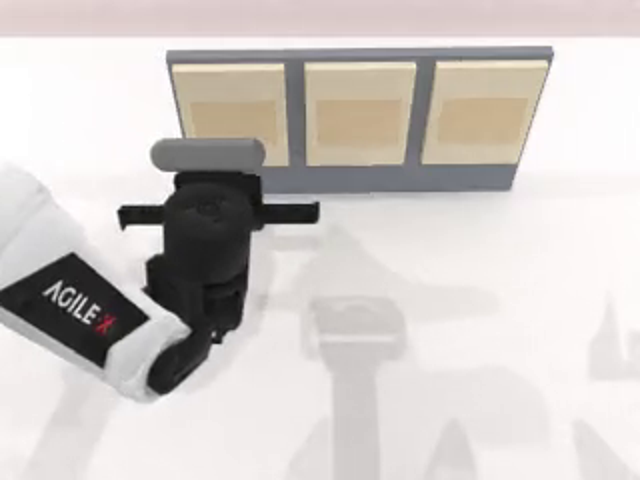} & \includegraphics[width=\appearancePanelWidth]{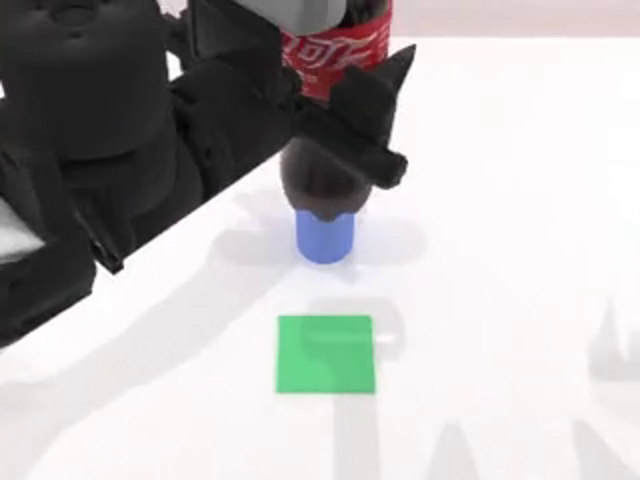} & \includegraphics[width=\appearancePanelWidth]{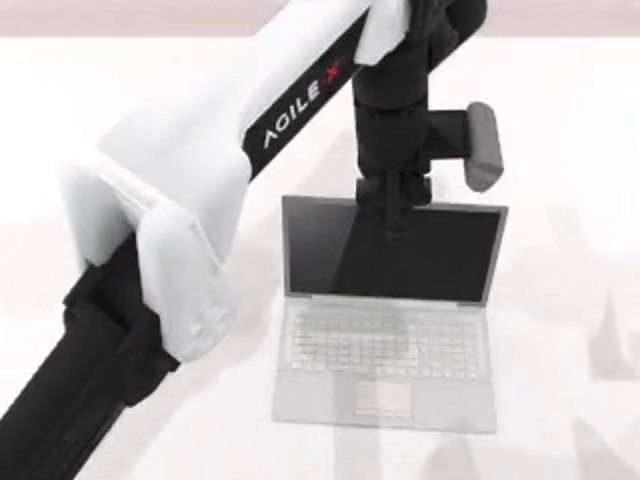} & \includegraphics[width=\appearancePanelWidth]{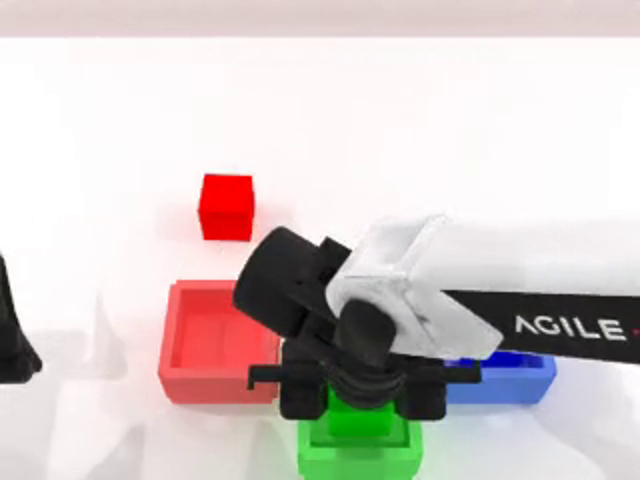} & \includegraphics[width=\appearancePanelWidth]{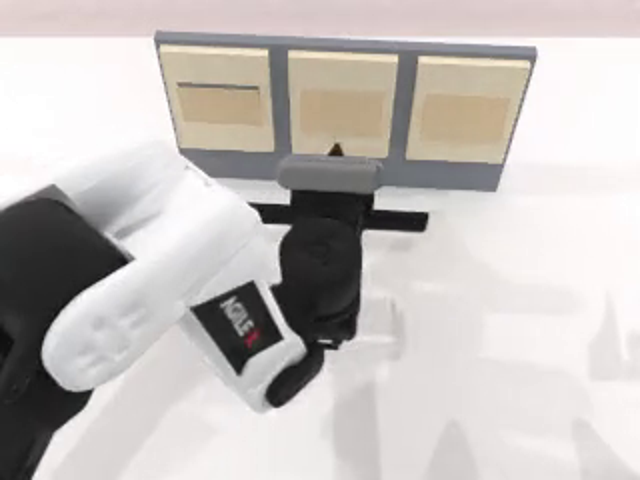} \\
\includegraphics[width=\appearancePanelWidth]{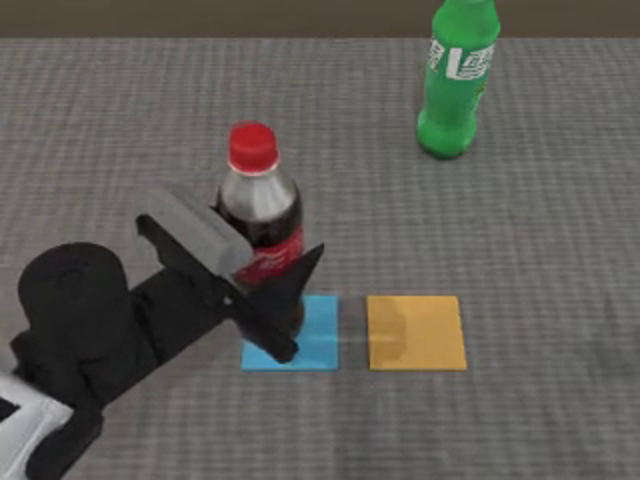} & \includegraphics[width=\appearancePanelWidth]{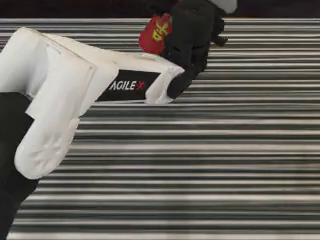} & \includegraphics[width=\appearancePanelWidth]{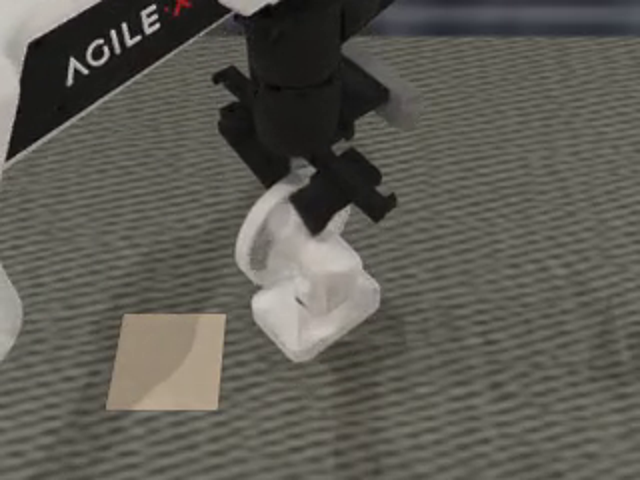} & \includegraphics[width=\appearancePanelWidth]{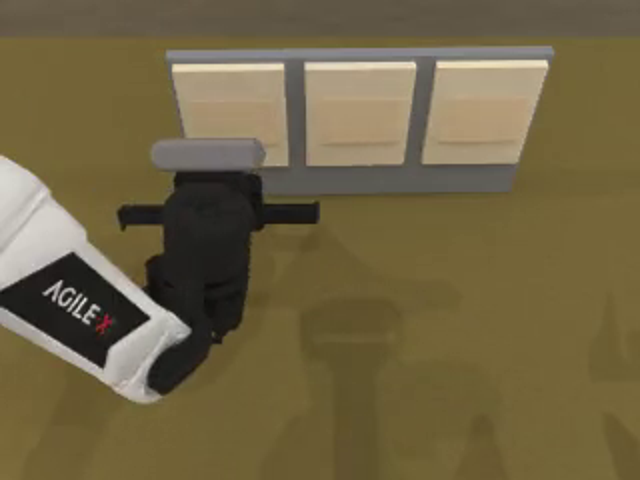} & \includegraphics[width=\appearancePanelWidth]{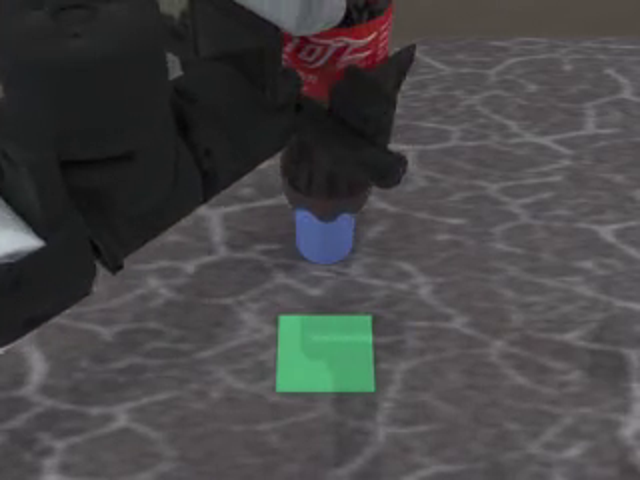} & \includegraphics[width=\appearancePanelWidth]{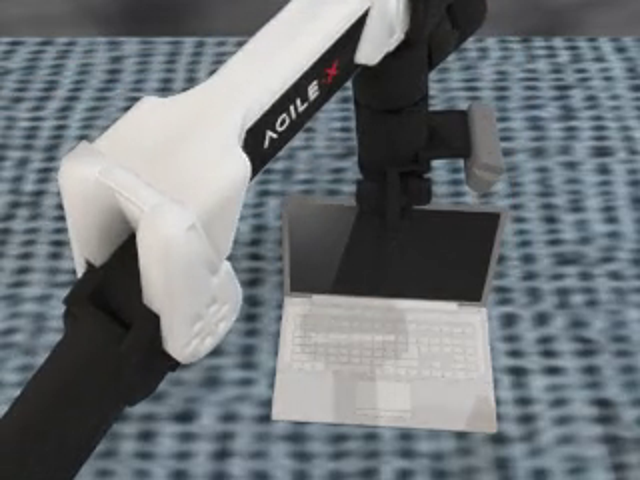} & \includegraphics[width=\appearancePanelWidth]{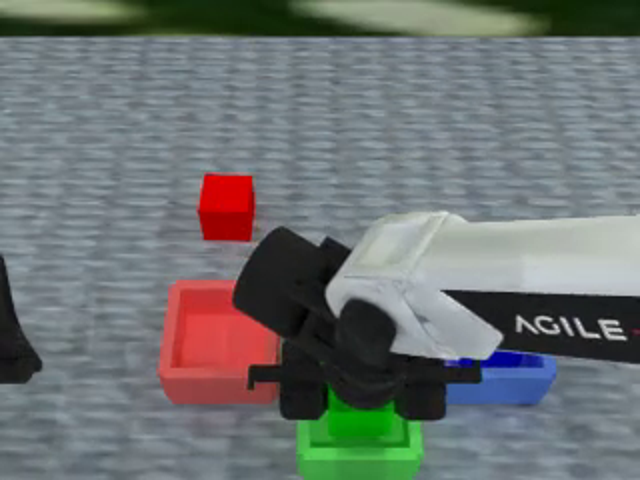} & \includegraphics[width=\appearancePanelWidth]{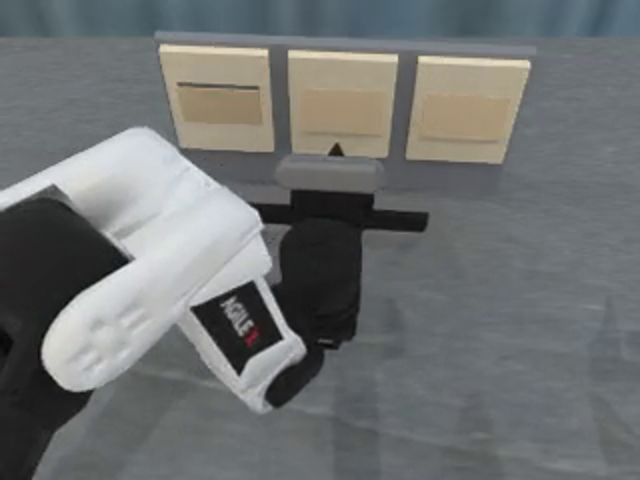} \\
\end{tabular}
\caption{Appearance conditions across the eight evaluation tasks. The top row shows clear appearance (\texttt{test\_clear}); the bottom row shows randomized appearance (\texttt{test\_random}). Each column pairs corresponding frames with the same instruction and physical state.}
\label{fig:appearance_pairs}
\end{figure*}

\paragraph{Data and evaluation conditions.}
We evaluate the eight tasks and instruction branches shown in Figure~\ref{fig:dataset_overview}. Dataset partitions are assigned by scene, with all sibling branches kept in the same split. Split sizes count \emph{episodes}: each episode is a demonstration of one instruction branch in one scene. We evaluate the same held-out episodes under clear appearance (\texttt{test\_clear}) and randomized appearance (\texttt{test\_random}). Motivated by evaluations of visual robustness under environmental perturbations~\cite{colosseum}, the latter reuses RoboTwin's appearance randomization~\cite{robotwin2} for textures, backgrounds, and lighting, with physical conditions fixed. Initial appearance and its sampling seed are shared across sibling branches. Appendix~\ref{sec:benchmarking_protocol} details the paired scenes and evaluation protocol.

\paragraph{Metrics.}
We report joint success and task success, corresponding to $J$ and $T$ in Sec.~\ref{sec:fidelity_definition}. For $N$ evaluated episodes, their rates are
\begin{equation}
    \mathrm{JR}=\frac{1}{N}\sum_{i=1}^{N}J_i,
    \qquad
    \mathrm{TR}=\frac{1}{N}\sum_{i=1}^{N}T_i.
\end{equation}
Each result cell presents $\mathrm{JR}/\mathrm{TR}$ as two percentages. Their difference, $\mathrm{TR}-\mathrm{JR}$, measures task completion without satisfying all applicable instruction requirements. We report results per task and average the eight task rates with equal weight. Both policy categories use the same task predicates and instruction checks; comparisons between methods are made within each category.

\subsection{Dataset Statistics and Expert Policy Generation}
\label{sec:expert_experiments}

Figure~\ref{fig:dataset_overview} shows the task coverage, same-state instruction branches, and simulator-grounded annotations.

The expert passed every instruction branch in 566 of 1,006 candidate scenes (56.3\%), yielding 4,242 demonstrations. The training, validation, and test splits contain 2,520, 861, and 861 demonstrations from 336, 115, and 115 scenes, respectively. Table~\ref{tab:collection_statistics} reports per-task collection statistics and expert refinement iterations.

\subsection{Qualitative Comparison of IR Acquisition}
\label{sec:ir_comparison}

Table~\ref{tab:ir_capabilities} compares the annotation fields available from the simulator and video-based pipelines. Our pipeline provides richer annotations, including ground-truth 3D point trajectories, contact positions and times, and metric 6D poses recorded directly from the simulator. Appendix~\ref{sec:ir_acquisition_supplement} provides further qualitative and quantitative analysis.

\subsection{Learning-based Policy Evaluation}
\label{sec:vla_experiments}

\begin{table*}[t]
    \centering
    \small
    \setlength{\tabcolsep}{3pt}
    \caption{Learning-based policy evaluation. Cells report JR/TR (\%); averages weight tasks equally. The evaluation protocol pairs clear and randomized episodes. \texttt{c2c} and \texttt{c2r} denote training on clear demonstrations and testing under clear and randomized appearance, respectively; \texttt{\_ir} denotes IR-derived command supervision with online VLM steering.}
    \label{tab:vla_results}
    \begin{tabular*}{\textwidth}{@{\extracolsep{\fill}}l*{10}{c}@{}}
        \toprule
        & \multicolumn{6}{c}{\texttt{test\_clear}} & \multicolumn{4}{c}{\texttt{test\_random}} \\
        \cmidrule(lr){2-7}\cmidrule(lr){8-11}
        & \multicolumn{2}{c}{Zero-shot transfer} & \multicolumn{2}{c}{$\pi_{0.5}$} & \multicolumn{2}{c}{OpenWAM} & \multicolumn{2}{c}{$\pi_{0.5}$} & \multicolumn{2}{c}{OpenWAM} \\
        \cmidrule(lr){2-3}\cmidrule(lr){4-5}\cmidrule(lr){6-7}\cmidrule(lr){8-9}\cmidrule(lr){10-11}
        Task & $\pi_{0.5}$ & OpenWAM & \texttt{c2c} & \texttt{c2c\_ir} & \texttt{c2c} & \texttt{c2c\_ir} & \texttt{c2r} & \texttt{c2r\_ir} & \texttt{c2r} & \texttt{c2r\_ir} \\
        \midrule
        Place bottle         & 0.0/1.0 & 0.0/0.0 & 76.9/87.5 & 53.8/53.8 & 100.0/100.0 & 29.8/30.8 & 64.4/64.4 & 37.5/42.3 & 67.3/74.0 & 31.7/35.6 \\
        Shake bottle & 0.0/0.0 & 0.0/45.2 & 1.9/99.0 & 1.9/89.4 & 0.0/100.0 & 0.0/70.2 & 0.0/87.5 & 0.0/89.4 & 0.0/95.2 & 0.0/69.2 \\
        Orient fan    & 0.0/0.0 & 5.8/5.8 & 50.0/50.0 & 10.6/10.6 & 13.5/13.5 & 3.8/3.8 & 36.5/36.5 & 7.7/7.7 & 1.9/1.9 & 1.0/1.0 \\
        Select switch          & 0.0/1.0 & 0.0/1.0 & 69.6/88.2 & 2.9/94.1 & 8.8/88.2 & 2.0/86.3 & 61.8/80.4 & 0.0/83.3 & 1.0/90.2 & 0.0/87.3 \\
        Transport via divider        & 0.0/0.0 & 0.0/0.0 & 45.6/100.0 & 51.8/100.0 & 62.3/92.1 & 39.5/99.1 & 57.9/94.7 & 36.0/94.7 & 44.7/83.3 & 28.9/75.4 \\
        Open laptop          & 2.6/2.6 & 10.3/18.8 & 16.2/17.9 & 10.3/13.7 & 11.1/12.0 & 6.0/8.5 & 12.0/16.2 & 17.1/21.4 & 7.7/13.7 & 8.5/12.8 \\
        Sort blocks in order     & 0.0/0.0 & 0.9/0.9 & 97.2/99.1 & 78.7/92.6 & 74.1/83.3 & 65.7/75.0 & 87.0/88.0 & 69.4/83.3 & 31.5/43.5 & 25.0/38.9 \\
        Press buttons in order   & 0.0/0.0 & 0.0/0.0 & 62.0/92.6 & 25.0/91.7 & 1.9/71.3 & 0.9/65.7 & 17.6/80.6 & 10.2/89.8 & 0.0/71.3 & 0.9/75.9 \\
        \midrule
        Average                  & 0.3/0.6 & 2.1/9.0 & 52.4/79.3 & 29.4/68.2 & 34.0/70.0 & 18.5/54.9 & 42.2/68.5 & 22.2/64.0 & 19.3/59.1 & 12.0/49.5 \\
        \bottomrule
    \end{tabular*}
\end{table*}

\paragraph{Settings.}
We evaluate two learning-based robot policies: the vision-language-action model $\pi_{0.5}$~\cite{pi05} and the world-action model OpenWAM~\cite{openwam_pretraining}. For experiments without IR, we follow the RoboTwin training and deployment setup~\cite{robotwin,robotwin2}, conditioning each policy on the task instruction. We evaluate both zero-shot transfer from RoboTwin-post-trained checkpoints and post-training on RoboInF's clear training split. Zero-shot transfer uses no additional training on RoboInF. Appendix~\ref{sec:policy_training} provides the training and inference details.

For experiments with IR, we follow Steerable Policies~\cite{steerable_policies}, using demonstration relabeling and online language steering. During training, simulator-grounded IR from RoboInF demonstrations is converted into diverse textual commands that condition each policy through its native text interface. During evaluation, an online Qwen3.8-Flash controller generates commands from the original task instruction, current visual observations, and episode history. Appendix~\ref{sec:steerable_ir} describes command annotation, policy training, and online steering.

\begin{table}[t]
    \centering
    \small
    \setlength{\tabcolsep}{5pt}
    \caption{Inference-only agents on \texttt{test\_clear}: JR/TR (\%) over 861 episodes each. Averages weight tasks equally.}
    \label{tab:agent_results}
    \begin{tabular}{@{}lcc@{}}
        \toprule
        Task & GPT-6 Astra & Qwen3.8-Flash \\
        \midrule
        Place bottle       & 57.7/57.7 & 59.6/59.6 \\
        Shake bottle & 0.0/64.4 & 0.0/39.4 \\
        Orient fan  & 0.0/0.0 & 1.0/1.0 \\
        Select switch        & 33.3/49.0 & 0.0/0.0 \\
        Transport via divider      & 10.5/11.4 & 3.5/15.8 \\
        Open laptop        & 26.5/26.5 & 1.7/1.7 \\
        Sort blocks in order   & 5.6/5.6 & 0.0/0.0 \\
        Press buttons in order & 7.4/16.7 & 0.0/0.0 \\
        \midrule
        Average                & 17.6/28.9 & 8.2/14.7 \\
        \bottomrule
    \end{tabular}
\end{table}

In Table~\ref{tab:vla_results}, \texttt{c2c} and \texttt{c2r} both use clear demonstrations for training and differ only in test appearance (clear or randomized, respectively); \texttt{\_ir} denotes Steerable Policies. Figure~\ref{fig:appearance_pairs} shows paired demonstrations under both conditions. For zero-shot transfer, we use \texttt{test\_clear}; for each post-trained checkpoint, we use all 861 episodes in both conditions.

\paragraph{Results.}
Table~\ref{tab:vla_results} shows that post-training on RoboInF improves average task success and joint success over zero-shot transfer. However, joint success remains below task success across both models and training configurations. This gap is especially evident in tasks such as Transport via divider and Shake bottle, where reaching the terminal goal does not establish whether the robot followed the requested route or motion pattern. The Steerable configurations also achieve lower average joint success than the task-instruction configurations under both appearance conditions, indicating that reliable execution of detailed instructions remains challenging with online language steering.

\subsection{Inference-only Agent Evaluation}
\label{sec:agent_experiments}

\paragraph{Settings.}
We evaluate GPT-6 Astra~\cite{gpt6astra} and Qwen3.8-Flash~\cite{qwen38flash} without parameter updates on all 861 test episodes of RoboInF. Both agents use the same decision interface and fixed RGB-D geometric skill executor, with the high-level model selected through its API. The model selects semantic skills, image targets, and skill parameters; the executor grounds these decisions and carries out robot motions. Appendix~\ref{sec:agent_implementation} describes the implementation.

\paragraph{Results.}
Under the shared skill interface (Table~\ref{tab:agent_results}), GPT-6 Astra achieves 28.9\% average task success but only 17.6\% joint success; for Qwen3.8-Flash the rates are 14.7\% and 8.2\%, respectively. The gap depends on the task: for Place bottle the two rates coincide for each agent, whereas for Shake bottle neither agent achieves joint success.

\section{Conclusion}
\label{sec:conclusion}

We introduced RoboInF, a benchmark for evaluating instruction fidelity in robotic manipulation. By pairing multiple expert-validated instruction branches from the same initial state with checks of both terminal outcomes and execution constraints, RoboInF supports controlled evaluation of goal selection, procedural adherence, and fine-grained specifications. Our multi-branch expert generation and simulator-grounded annotation pipeline produces RoboInF-4K, pairing 4,242 demonstrations with structured intermediate representations of geometry, motion, and contact. Evaluations of learned manipulation policies and inference-only agents reveal persistent gaps between task completion and faithful execution. Post-training improves performance, but requirements on how a task is executed remain challenging. These findings motivate evaluating instruction fidelity alongside task success and developing policies that respond more reliably to detailed behavioral requirements. Extending the benchmark to broader task families and real-world settings is a promising direction for future work.

\paragraph{Limitations.}
Due to equipment and time constraints, RoboInF-4K is limited to 4,242 demonstrations across eight tasks; collecting more demonstrations across a wider range of tasks would broaden its coverage.

\FloatBarrier
{\small
\bibliographystyle{ieeenat_fullname}
\bibliography{main}
}
\clearpage
\onecolumn
\appendix
\section*{Appendix}
\pdfbookmark[0]{Appendix}{appendix-start}
\section*{Contents}
\begingroup
\setcounter{tocdepth}{2}
\makeatletter\@starttoc{atc}\makeatother
\endgroup
\twocolumn
\let\originaladdcontentsline\addcontentsline
\renewcommand{\addcontentsline}[3]{%
  \ifstrequal{#1}{toc}{\originaladdcontentsline{atc}{#2}{#3}}{\originaladdcontentsline{#1}{#2}{#3}}}
\begingroup
\raggedbottom
\section{Qualitative and Quantitative Analysis of IR Annotation Methods}
\label{sec:ir_acquisition_supplement}

\subsection{IR fields provided by the evaluated video pipelines}
\label{sec:astra_ir_audit}
We evaluate the four RGB annotation pipelines in Table~\ref{tab:ir_capabilities} of the main paper on one complete test scene per task. The eight scenes contain 63 branch executions. We score 32 uniformly sampled frames per video (2,016 frames total) against simulator references. All four pipelines operate without branch targets, simulator labels, depth, or camera calibration.

GPT-6 Astra~\cite{gpt6astra} receives the 32 sampled RGB frames in temporal order. A fixed prompt names the task objects and landmarks and allows the model to abstain. The remaining pipelines process the full head-camera video.

Following the Grounded-SAM-2 (GSAM) tracking demo~\cite{grounded_sam2}, we detect objects in the first frame with Grounding DINO~\cite{grounding_dino} and propagate their masks with SAM~2~\cite{sam2}. For repeated categories, we retain the expected number of highest-scoring detections and assign object identities from left to right. If the first frame yields too few detections, we omit that category for the video.

CoTracker3~\cite{cotracker3} adds point trajectories to the GSAM boxes and masks. An annotator initializes each landmark in the earliest frame where it can be identified and marks uncertain landmarks as such. CoTracker3 then tracks the points forward and backward without manual correction.

For RoboInter-Tools~\cite{robointer}, the same annotator supplies foreground and background clicks, reviews the propagated masks, and revises the prompts as needed. For each manipulated object, the annotator also marks the first visible contact during its manipulation stage.

\begin{figure*}[t]
\centering
\includegraphics[width=\textwidth]{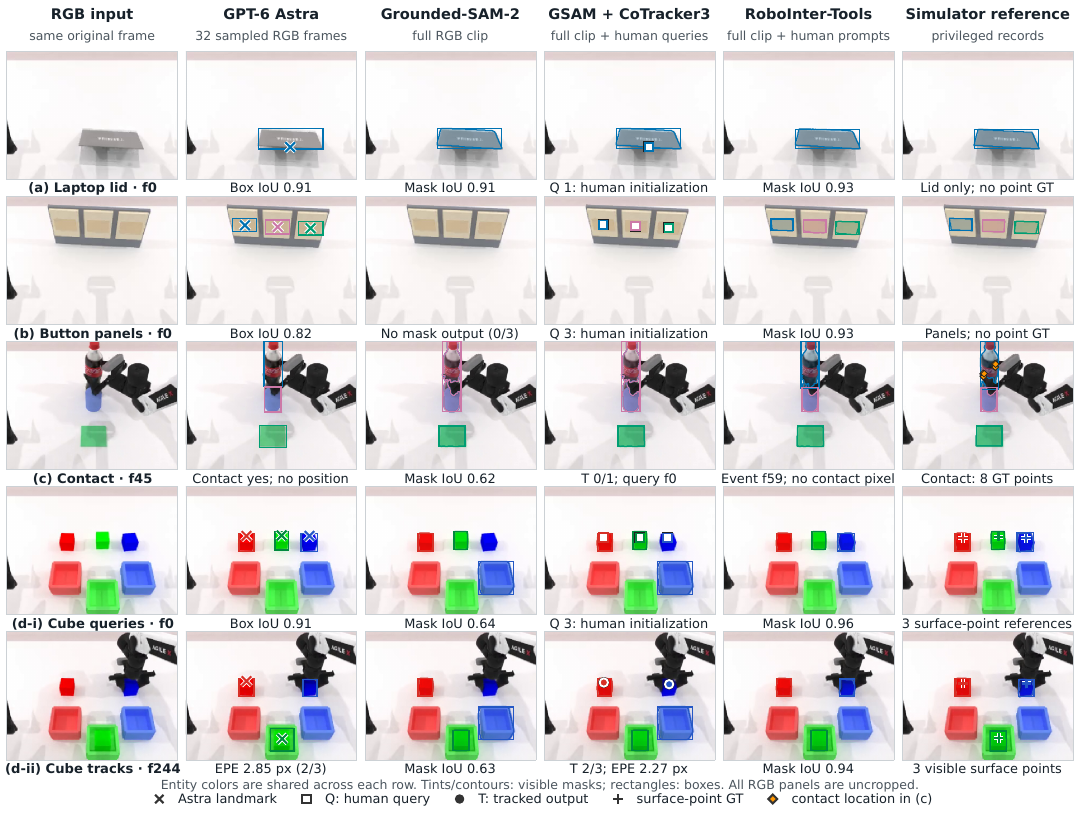}
\caption{RGB observations, pipeline outputs and simulator references for the same frames. (a) Both automatic methods localize the laptop lid. (b) Grounding DINO detects the panel as a whole. (c) The simulator records physical contact (orange diamonds), but Astra supplies no contact coordinate. (d) Human-selected cube landmarks (squares) initialize CoTracker; circles show its later estimates. Crosses mark Astra landmarks, plus signs mark reference landmarks, tints show masks, and rectangles show boxes. Frame IoUs average over returned objects with valid references.}
\label{fig:ir_acquisition_cases}
\end{figure*}

\begin{figure*}[t]
\centering
\includegraphics[width=\textwidth]{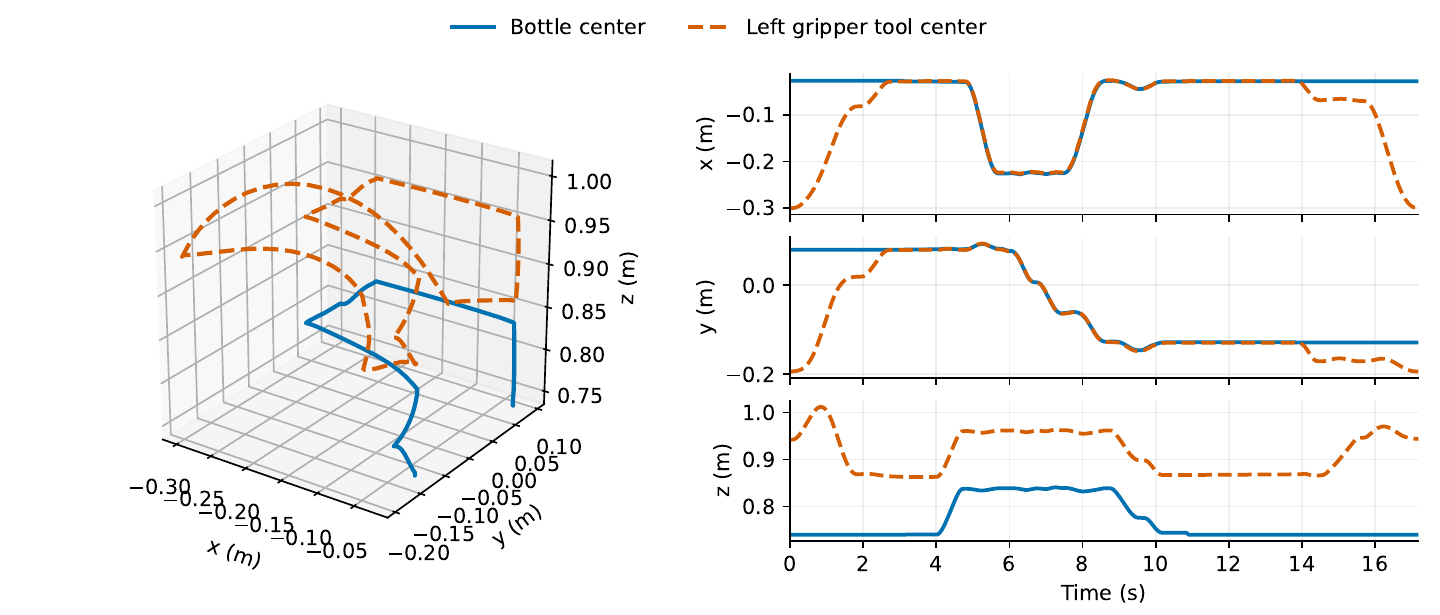}
\caption{The left gripper carries a bottle around the left side of the divider. Left: trajectories of the bottle center (solid blue) and the gripper tool center (dashed orange) in world coordinates. Right: the same points' $x$, $y$ and $z$ coordinates over time, in meters.}
\label{fig:ir_metric_fields}
\end{figure*}

\subsection{Localization and physical correspondence}
The simulator supplies 4,881 nonempty reference masks and 4,861 positive-area reference boxes across the sampled frames. We compute mean intersection over union (IoU) on returned annotations. Coverage and the fraction reaching IoU$\geq0.5$ use all eligible references as the denominator. We average scores within each execution, then across branches within each scene, and finally across tasks with equal weight.

\begin{table}[!htb]
\centering\small
\caption{Object localization and human annotation time. GSAM returns no identified buttons: its mean IoU covers seven tasks, whereas coverage and threshold scores include all eight. Time is approximate active time per video from browser logs; sessions can overlap, and model waits are excluded. A dash marks an unsupported field.}
\label{tab:ir_acquisition_localization}
\setlength{\tabcolsep}{2.3pt}
\begin{tabular}{@{}lrrrr@{}}
\toprule
Pipeline & \shortstack{Mean IoU\\(returned)} & \shortstack{Coverage\\(\%)} & \shortstack{IoU$\geq$.5\\(\% eligible)} & \shortstack{Human time\\(s/episode)}\\
\midrule
\multicolumn{5}{@{}l}{\emph{Boxes}}\\
Astra & .840 & 98.9 & 97.1 & 0\\
GSAM & .665 & 84.4 & 58.5 & 0\\
\quad + CoTracker3 & .665 & 84.4 & 58.5 & 26\\
RoboInter & .909 & 98.5 & 95.2 & 150\\
\midrule
\multicolumn{5}{@{}l}{\emph{Masks}}\\
Astra & -- & -- & -- & 0\\
GSAM & .661 & 83.8 & 59.2 & 0\\
\quad + CoTracker3 & .661 & 83.8 & 59.2 & 26\\
RoboInter & .904 & 98.0 & 95.9 & 150\\
\bottomrule
\end{tabular}
\end{table}

RoboInter has the highest mean IoU on returned boxes (.909), while Astra reaches IoU$\geq0.5$ on the largest fraction of eligible boxes (97.1\%; Table~\ref{tab:ir_acquisition_localization}). In the button scene, Grounding DINO detects the panel rather than its individual buttons (Fig.~\ref{fig:ir_acquisition_cases}b).

We evaluate landmarks on the block task, where the cube top-face centers have matching simulator projections. In the other tasks, simulator link origins and requested surface landmarks refer to different physical points. Table~\ref{tab:block_landmarks} compares Astra's direct estimates with tracks initialized by a human annotator.

\begin{table}[!htb]
\centering\small
\caption{Localization of cube top-face centers. Eligible instances are visible landmarks in sampled frames; for CoTracker3 we exclude each point's manual initialization frame. Mean error is computed only for returned points and averaged by execution.}
\label{tab:block_landmarks}
\begin{tabular}{@{}lrrr@{}}
\toprule
Method & Returned & Eligible & Mean error (px)\\
\midrule
Astra & 453 & 518 & 1.75\\
CoTracker3 & 420 & 500 & 2.55\\
\bottomrule
\end{tabular}
\end{table}

\subsection{Contact annotations}
Astra classifies contact correctly on 98.3\% of frames for which it gives a yes/no answer. Across all eligible frames, including abstentions, it gives a correct answer on 70.7\% (Table~\ref{tab:astra_ir_summary}).

Across the 1,129 sampled frames with simulator-recorded contact, Astra returns a 2D contact point in 113 (10.0\% pooled across tasks; Fig.~\ref{fig:ir_acquisition_cases}c). This denominator includes 73 frames in which all contacts project outside the image. We score each returned point against the nearest projected contact, using contacts on the named entity when one is specified. Four tasks have scorable coordinates; their mean error is 4.02 pixels.

\noindent\begin{minipage}{\columnwidth}
\centering\small
\captionsetup{hypcap=false}\captionof{table}{Task-averaged contact scores. The evaluation unit is a sampled frame for Astra and an object--execution pair for RoboInter.}
\label{tab:astra_ir_summary}
\setlength{\tabcolsep}{2pt}
\begin{tabular}{@{}lr@{}}
\toprule
Measurement & Result\\
\midrule
\multicolumn{2}{@{}l}{\emph{Astra: states and image points per frame}}\\
Explicit yes/no answers / all frames & 72.0\%\\
Accuracy conditional on a yes/no answer & 98.3\%\\
Correct yes/no output / all eligible frames & 70.7\%\\
Position returned / frames with physical contact & 11.8\%\\
Nearest projected contact error, when scored & 4.02 px\\
\midrule
\multicolumn{2}{@{}l}{\emph{RoboInter: first event per object and execution}}\\
Event returned / pairs with recorded contact & 96.9\%\\
Mean absolute offset from first recorded contact & 1.325 s\\
\bottomrule
\end{tabular}
\end{minipage}\par

RoboInter returns a contact event for 95 of the 99 object--execution pairs with recorded contact and one event for a pair without recorded contact. For matched pairs, the mean absolute offset is 1.325 seconds. The visual annotation marks contact during manipulation, whereas the simulator reference is the first physical contact associated with a camera frame and can precede manipulation.

\subsection{Human input and repeatability}
Across the 63 executions, the annotator supplied 99 landmark queries for CoTracker3 and 243 segmentation prompts for RoboInter, in addition to reviewing the RoboInter masks and marking contact frames.

\paragraph{Within-annotator repeatability.}
The annotator repeated two predetermined executions per task (16 total) after at least 24 hours, without seeing the original annotations or simulator references.
Of 24 landmarks identified in both rounds, 22 were selected in the same frame. For these 22 pairs, the mean click distance was 1.31 pixels, with equal weight given to the five contributing tasks. For 24 contact events matched by object and chronological order, the mean absolute difference between the two visual annotations was 0.052 seconds across eight tasks.

\newpage
\subsection{Metric pose availability}
Our simulator records object poses in world coordinates for all 4,960 object observations in the evaluation, whereas none of the tested RGB pipelines outputs metric 6D poses. Figure~\ref{fig:ir_metric_fields} shows the coordinated motion of the gripper and bottle as they pass around the divider.
\clearpage
\endgroup

\onecolumn
\section{Task Design Details}
\label{sec:task_design_details}

Table~\ref{tab:task_design_details} pairs each instruction with its terminal goal and execution constraints. For example, a bottle can finish in the requested holding pose even when the robot performs fewer than three shake cycles. Such a rollout can satisfy $T$ but fails $J$.

\begin{center}
    \centering
    \fontsize{9}{10.5}\selectfont
    \setlength{\tabcolsep}{4pt}
    \renewcommand{\arraystretch}{1.08}
    \captionsetup{hypcap=false}\captionof{table}{Canonical instructions and episode-level success criteria. Task success ($T$) requires the terminal goal; joint success ($J$) also requires the conditions in the last column. Braces enclose instruction slots, with alternatives separated by slashes.}
    \label{tab:task_design_details}
    \begin{tabular}{@{}p{0.12\textwidth}p{0.38\textwidth}p{0.20\textwidth}p{0.25\textwidth}@{}}
        \toprule
        \textbf{Task} & \textbf{Canonical instruction} & \textbf{Terminal goal ($T$)} & \textbf{Additional conditions for $J$} \\
        \midrule
        \raggedright Place bottle\par Fig.~\ref{fig:supp-01_place_the_bottle}
        & \raggedright Move \{bottle\_left/\allowbreak bottle\_right\} to \{goal\_left/\allowbreak goal\_right\} using the \{left/\allowbreak right\} arm.
        & \raggedright Target bottle center inside the requested goal; target bottle upright; active gripper open.
        & \raggedright Use the requested arm; keep the non-target bottle undisturbed and upright. \tabularnewline
        \midrule
        \raggedright Shake bottle\par Fig.~\ref{fig:supp-02_shake_bottle}
        & \raggedright Use the \{left/\allowbreak right\} arm to hold the bottle \{horizontally/\allowbreak vertically\} and shake it \{left and right three times/\allowbreak with three small wrist rotations around the horizontal grasp axis\}.
        & \raggedright Bottle contacted and held, with the gripper closed at completion; requested final holding pose.
        & \raggedright Hold and shake with the requested arm; maintain the requested holding pose during shaking; execute exactly three cycles of the requested motion: left--right shaking or small wrist rotations around the horizontal grasp axis. \tabularnewline
        \midrule
        \raggedright Orient fan\par Fig.~\ref{fig:supp-03_place_fan_orientation}
        & \raggedright Use the \{left/\allowbreak right\} arm to place the fan facing \{front/\allowbreak front right/\allowbreak right/\allowbreak back right/\allowbreak back/\allowbreak back left/\allowbreak left/\allowbreak front left\} on the pad.
        & \raggedright Fan at the pad with the requested orientation; fan upright; both grippers open.
        & \raggedright Use the requested, expert-resolved arm. \tabularnewline
        \midrule
        \raggedright Select switch\par Fig.~\ref{fig:supp-04_turn_switch_row}
        & \raggedright Turn on the \{left/\allowbreak middle/\allowbreak right\} switch using the \{left/\allowbreak right\} arm.
        & \raggedright Target switch ON; released and stable.
        & \raggedright Operate the target switch using the requested arm while keeping the other switches undisturbed throughout execution. \tabularnewline
        \midrule
        \raggedright Transport via divider\par Fig.~\ref{fig:supp-05_place_via_divider}
        & \raggedright Move the beverage bottle to the target zone using the \{left/\allowbreak right\} arm, \{passing around the left side of/\allowbreak passing around the right side of/\allowbreak passing directly over\} the divider.
        & \raggedright Bottle at the target, upright and stable; both grippers open.
        & \raggedright Use the requested arm; measured bottle trajectory follows the requested route relative to the divider. \tabularnewline
        \midrule
        \raggedright Open laptop\par Fig.~\ref{fig:supp-06_open_the_laptop}
        & \raggedright Open the laptop to \{45/\allowbreak 90/\allowbreak 135\} degrees by contacting the \{left/\allowbreak center/\allowbreak right\} region of the lid.
        & \raggedright Requested opening angle reached; lid stable; joint within legal limits.
        & \raggedright Contact the requested lid region. \tabularnewline
        \midrule
        \raggedright Sort blocks in order\par Fig.~\ref{fig:supp-07_sort_blocks_in_order}
        & \raggedright Sort the blocks in this order: \{first\}, then \{second\}, then \{third\}.
        & \raggedright All blocks in matching bins and on the table; both grippers open; final configuration stable.
        & \raggedright Measured placement order equals the requested color permutation. \tabularnewline
        \midrule
        \raggedright Press buttons in order\par Fig.~\ref{fig:supp-08_turn_switches_in_order}
        & \raggedright Using the \{left/\allowbreak right\} arm, press the \{left/\allowbreak middle/\allowbreak right\}, \{left/\allowbreak middle/\allowbreak right\}, and \{left/\allowbreak middle/\allowbreak right\} buttons in that order.
        & \raggedright All three switches ON at completion.
        & \raggedright Measured activation sequence equals the requested button permutation; all presses use the requested arm. \tabularnewline
        \bottomrule
    \end{tabular}
    \par\smallskip
    \begin{minipage}{\textwidth}
        \fontsize{9}{10.5}\selectfont
        The block and button orders are permutations of the three colors (red, green, blue) and three positions (left, middle, right), respectively. Each fan scene has eight orientation branches; the expert determines the arm for each branch. Laptop and block-sorting instructions leave the arm unspecified. All switches start OFF in Select switch.
    \end{minipage}
\end{center}

Each row in the task figures shows selected stages of one expert execution in temporal order. Paired in-distribution (ID) examples start from the same initial scene. The bottle pair varies arm, object, and destination; the laptop pair varies contact region and angle.

Figure~\ref{fig:supp-09_id_ood_comparison} shows the corresponding clear and randomized views for a divider traversal.

\twocolumn
\begingroup
\makeatletter
\setlength{\@dblfptop}{0pt}
\makeatother
\begin{figure*}[t]
    \centering
    \includegraphics[width=\textwidth]{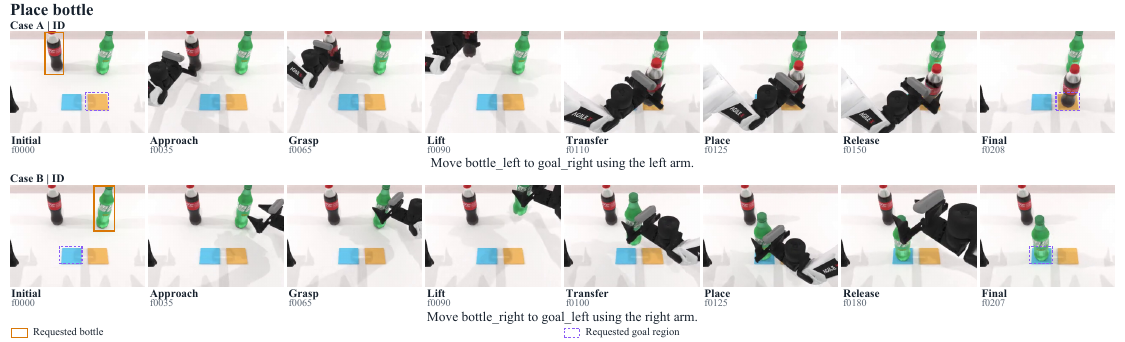}
    \caption{Place bottle. From the same initial scene, Case A moves bottle\_left to goal\_right with the left arm; Case B moves bottle\_right to goal\_left with the right arm. Orange outlines mark the requested bottle; purple dashed outlines mark the requested goal.}
    \label{fig:supp-01_place_the_bottle}
\end{figure*}

\begin{figure*}[t]
    \centering
    \includegraphics[width=\textwidth]{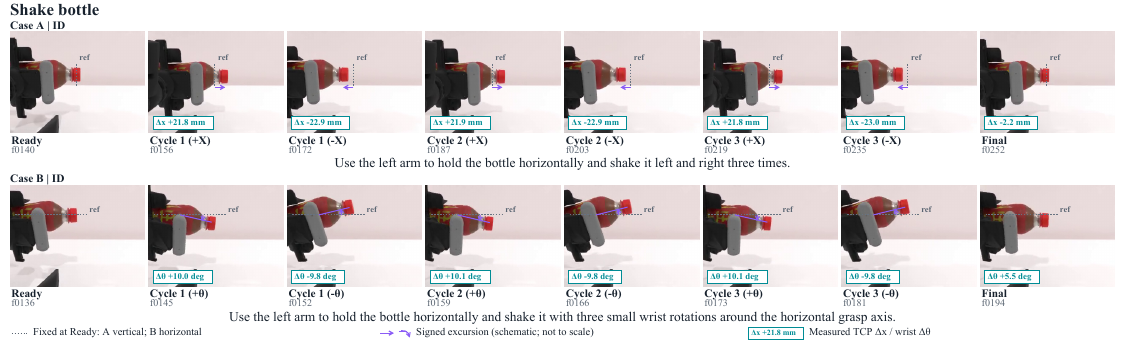}
    \caption{Shake bottle. Same initial scene, left arm and horizontal holding pose; only lateral translation versus wrist rotation changes. Six extrema show three cycles. Case A uses a vertical cap reference; Case B uses a fixed horizontal reference at the initial Ready bottle height. Purple arrows are schematic; teal values are measured relative to Ready.}
    \label{fig:supp-02_shake_bottle}
\end{figure*}

\begin{figure*}[t]
    \centering
    \includegraphics[width=\textwidth]{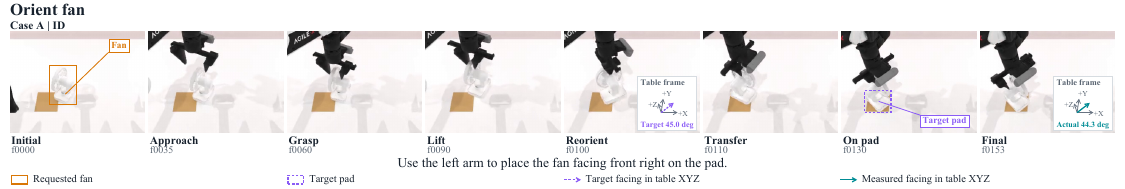}
    \caption{Orient fan. Object, target pad and facing annotations appear in separate panels. The calibrated table-frame insets show target and measured facing separately: +X right, +Y front and +Z up. Facing angles are measured from +X toward +Y.}
    \label{fig:supp-03_place_fan_orientation}
\end{figure*}

\begin{figure*}[t]
    \centering
    \includegraphics[width=\textwidth]{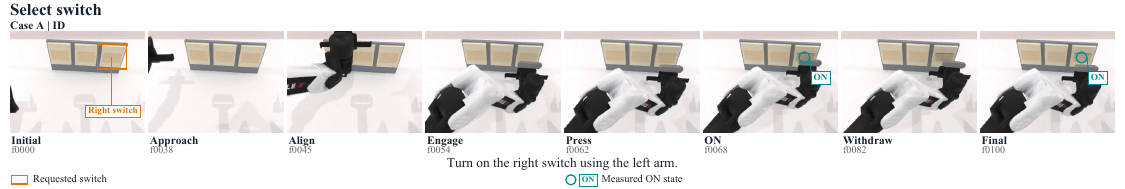}
    \caption{Select switch. The left arm activates only the right switch. The other two switches remain OFF. Orange marks the requested switch; the teal ring marks its measured ON transition (normalized joint progress $\geq$ 0.95).}
    \label{fig:supp-04_turn_switch_row}
\end{figure*}

\begin{figure*}[t]
    \centering
    \includegraphics[width=\textwidth]{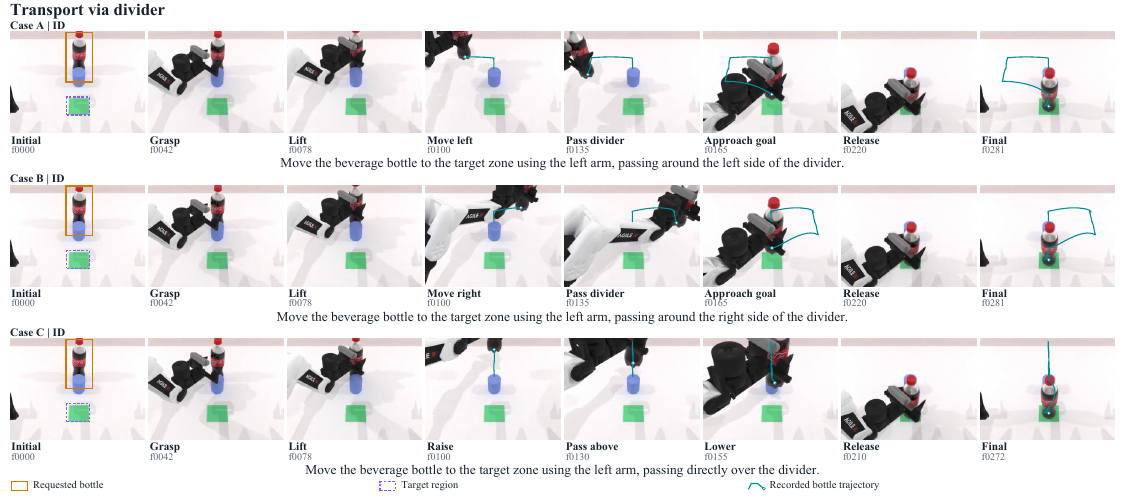}
    \caption{Transport via divider. The same scene, left arm and target are used for left-around, right-around and directly-over routes. Teal curves are recorded bottle-root trajectories projected into the camera.}
    \label{fig:supp-05_place_via_divider}
\end{figure*}

\begin{figure*}[t]
    \centering
    \includegraphics[width=\textwidth]{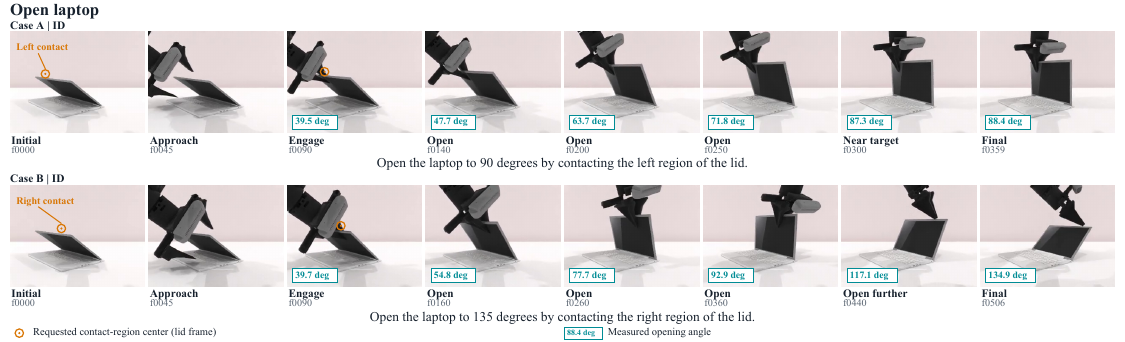}
    \caption{Open laptop. Case A requests left contact and 90 degrees; Case B requests right contact and 135 degrees from the same initial scene. Orange circles locate the requested contact-region centers in the lid frame; teal labels show measured opening angles.}
    \label{fig:supp-06_open_the_laptop}
\end{figure*}

\begin{figure*}[t]
    \centering
    \includegraphics[width=\textwidth]{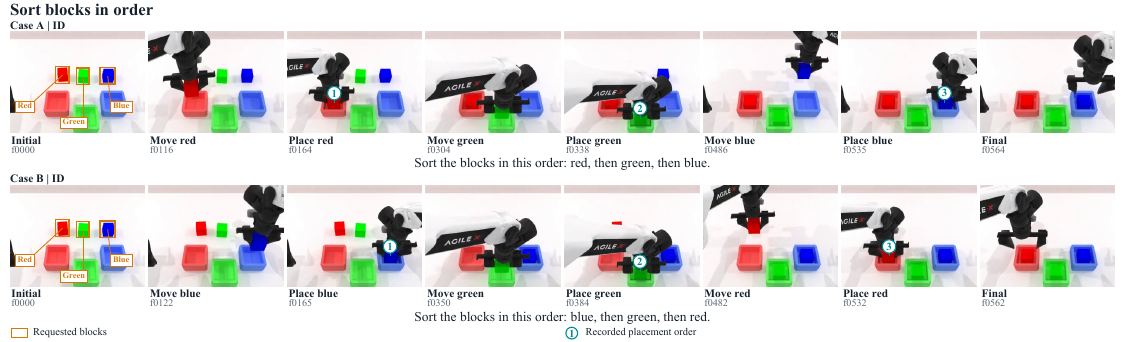}
    \caption{Sort blocks in order. Same initial state; only the requested sorting order changes. Both branches reach matching color bins. Numbered markers identify the successive placements; arrivals are located from recorded object positions.}
    \label{fig:supp-07_sort_blocks_in_order}
\end{figure*}

\begin{figure*}[t]
    \centering
    \includegraphics[width=\textwidth]{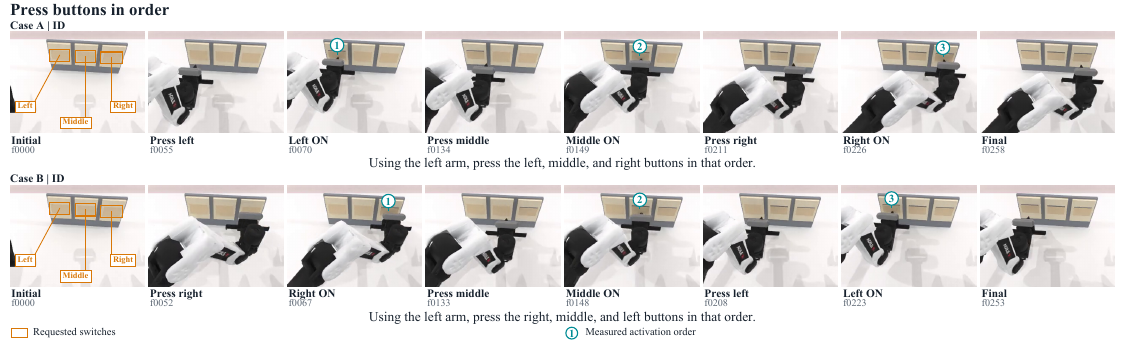}
    \caption{Press buttons in order. Same initial state and left arm; only the requested activation order changes. Numbered teal markers identify successive ON transitions.}
    \label{fig:supp-08_turn_switches_in_order}
    \par\medskip
    \includegraphics[width=\textwidth]{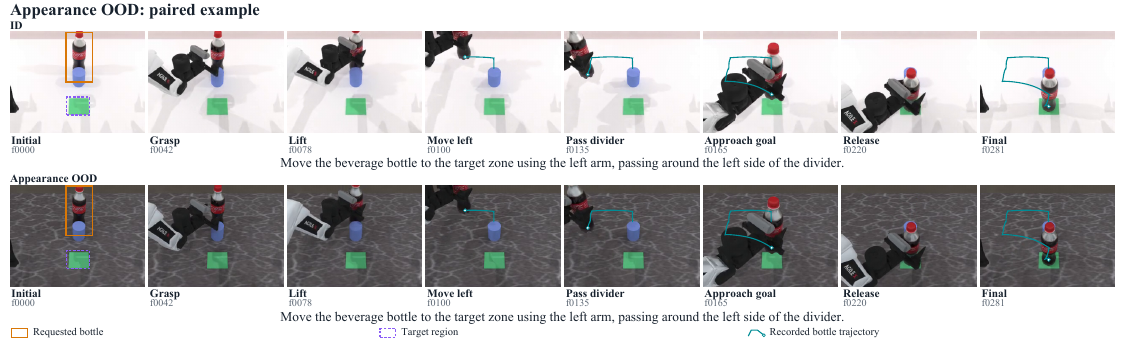}
    \caption{Paired expert demonstrations under clear (top) and randomized (bottom) appearance, with the same instruction, initial physical state, and frame indices.}
    \label{fig:supp-09_id_ood_comparison}
\end{figure*}

\FloatBarrier
\endgroup

\section{Benchmarking Protocol}
\label{sec:benchmarking_protocol}

\subsection{Dataset Collection and Expert Refinement}
\label{sec:collection_details}

A candidate scene is eligible when the expert satisfies $J$ on every instruction branch. Collection stops at the first failed branch and discards that scene. Table~\ref{tab:collection_statistics} of the main paper reports the fraction of candidate scenes retained for each task.

Refinement uses the development scenes and acceptance criterion in Sec.~\ref{sec:data_engine} of the main paper. Iteration 0 denotes the initial program; each revision counts as one iteration. The collection rates use the final expert on separately sampled candidate scenes.

\subsection{Evaluation Conditions and Scoring}

\paragraph{Scene reconstruction and pairing.}
The test split contains 861 instruction branches across eight tasks, each evaluated under clear and randomized appearance. We reconstruct and verify the initial object and robot states. Sibling branches share the same physical scene and camera configuration; each clear/random pair also shares its instruction and action budget.

\paragraph{Appearance randomization.}
We reuse RoboTwin's visual domain-randomization routines~\cite{robotwin2} to vary table and wall textures, backgrounds, and the RGB colors of directional and point lights. Textures are sampled from RoboTwin's unseen evaluation pool. Each surface has a 0.02 probability of using a clean background, and dynamic lighting is enabled with probability 0.02. A dedicated appearance random stream is independent of scene initialization. Sibling instruction branches share the initial appearance specification and seed; dynamic lighting subsequently follows the simulator's rendering updates.

The paired scenes have identical geometry, object and robot configurations, physical properties, and camera calibration. Neither condition adds clutter. Both use ray tracing with 32 samples per pixel, path depth 8, and OIDN denoising.

\paragraph{Execution budgets.}
Both task-only and Steerable policies replan every 20 actions and run to the task-specific limit below. After the last action, the simulator advances 250 passive physics steps before final scoring.
\begin{center}
\small
\begin{tabular}{@{}rp{0.72\columnwidth}@{}}
\toprule
Actions & Tasks \\
\midrule
400 & Place bottle, Orient fan, Select switch, Transport via divider \\
600 & Press buttons in order \\
700 & Shake bottle, Open laptop \\
1,000 & Sort blocks in order \\
\bottomrule
\end{tabular}
\end{center}

\paragraph{Scoring and failures.}
Checkers evaluate simulator states and execution records; their outputs are unavailable to the policy. We report JR and TR per task and average the eight task rates equally. Only infrastructure interruptions trigger a restart, using the same scene and method configuration.

\section{Task-only Policy Training and Evaluation}
\label{sec:policy_training}

In the task-only conditions of Table~\ref{tab:vla_results} (\texttt{c2c} and \texttt{c2r}), both policies receive the original task instruction throughout training and evaluation.

\paragraph{RoboTwin configuration and data.}
We retain each policy's RoboTwin embodiment, camera ordering, native observation and action interfaces, and training objective. We initialize $\pi_{0.5}$ from the RoboTwin cotrained checkpoint at step 29,999 and OpenWAM from \texttt{OpenWAM-Alpha-Sim-RoboTwin-Full} at step 118,655. Each model is then trained for 30,000 updates on the 2,520 clear training episodes of RoboInF-4K. Sampling balances tasks by their available training-window counts. Normalization statistics are recomputed from the training split only, using the native quantile normalization for $\pi_{0.5}$ and min--max normalization for OpenWAM.

\paragraph{Observation and action representations.}
$\pi_{0.5}$ receives head and left/right wrist RGB images, each resized and padded to $224\times224$, together with the 14D joint-and-gripper state. It learns joint deltas and absolute gripper targets, padded to the native 32D action space. OpenWAM assembles the three views into a $384\times320$ canvas: the head view occupies the upper $256\times320$ region, and two $128\times160$ wrist views occupy the lower region. Its 20D dual-arm end-effector representation contains position, 6D rotation encoding, and gripper state, mapped to the native 80D action space. During inference, predicted 6D rotations are converted to the orientation format used by the controller. We retain the RoboTwin instruction template for OpenWAM.

\begin{table}[t]
    \centering
    \small
    \setlength{\tabcolsep}{3pt}
    \renewcommand{\arraystretch}{1.08}
    \caption{Task-only post-training settings. Both models use eight GPUs, global batch 8, seed 0, eight data workers, and 30,000 optimizer updates.}
    \label{tab:policy_training}
    \begin{tabular}{@{}p{0.36\columnwidth}p{0.27\columnwidth}p{0.30\columnwidth}@{}}
        \toprule
        Setting & $\pi_{0.5}$ & OpenWAM \\
        \midrule
        Action horizon & 50 & 32 \\
        Window start stride & 2 & 1 \\
        Video stride & -- & 4 \\
        Peak learning rate & $10^{-5}$ & $10^{-4}$ \\
        Warmup updates & 1,000 & 1,500 \\
        Cosine decay endpoint & 20,000 & 30,000 \\
        Minimum learning rate & $10^{-6}$ & $10^{-6}$ \\
        AdamW $\beta_1,\beta_2$ & $0.9,0.95$ & $0.9,0.95$ \\
        AdamW $\epsilon$ & $10^{-8}$ & $10^{-8}$ \\
        Weight decay & $10^{-10}$ & $0.01$ \\
        Gradient norm clip & 1.0 & 1.0 \\
        EMA decay & 0.99 & -- \\
        Color jitter & Enabled & Disabled \\
        Checkpoint interval & 2,000 & 2,000 \\
        \bottomrule
    \end{tabular}
\end{table}

\paragraph{Optimization.}
We update all $\pi_{0.5}$ parameters with the action flow-matching loss. Its learning rate remains at the minimum after 20K updates. OpenWAM trains its video and action denoisers with equally weighted flow-matching losses and frozen text and visual encoders. Neither policy uses geometric cropping or rotation; OpenWAM also disables mirroring. Table~\ref{tab:policy_training} gives the optimizer and color augmentation settings.

\paragraph{Inference and full-test evaluation.}
For each model, we use the final 30K checkpoint and training-split normalization in both appearance conditions (Appendix~\ref{sec:benchmarking_protocol}). $\pi_{0.5}$ predicts 50 actions in 10 sampling steps. OpenWAM predicts 32 actions in 10 synchronous denoising steps with guidance scale 1; approximate denoiser caching and video decoding are disabled. Each policy executes the first 20 actions before replanning with the original instruction.

\FloatBarrier

\section{Steerable IR Annotation, Training, and Evaluation}
\label{sec:steerable_ir}

\begin{figure*}[t]
    \centering
    \includegraphics[width=\textwidth]{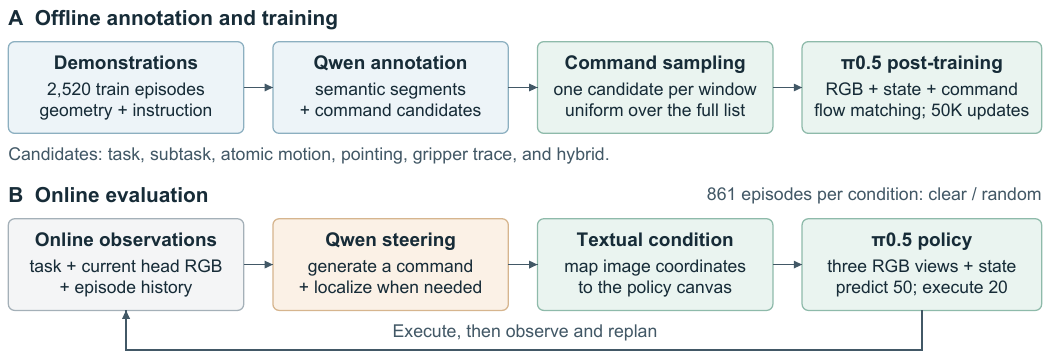}
    \caption{\textbf{Steerable $\pi_{0.5}$ training and evaluation.} Offline, measured demonstration geometry supports semantic segmentation and diverse command annotation. A sampled command conditions the native action-learning objective. Online, Qwen generates commands from the original task, current head image, and episode history; optional localization supplies image coordinates. The trained policy executes 20 actions before receiving a new command.}
    \label{fig:steerable_ir_workflow}
\end{figure*}

Following Steerable Policies~\cite{steerable_policies}, we relabel demonstrations with language commands for training and generate commands online during evaluation. Qwen3.8-Flash provides both the offline annotations and online commands. Figure~\ref{fig:steerable_ir_workflow} illustrates the procedure for $\pi_{0.5}$; OpenWAM uses the same annotation and steering procedure.

\subsection{Offline Command Annotation}

\paragraph{Geometric evidence.}
For each of the 2,520 training demonstrations, we extract gripper aperture, the midpoint between each gripper's fingertips, object poses, and visible object boxes from the simulator IR. Camera calibration maps each gripper midpoint into the head image; an occluded midpoint remains valid if its projection lies within the image. Object points are the centers of valid visible boxes. We express both kinds of point on a 0--255 grid with the origin at the upper left and mask invalid or out-of-view points.

\paragraph{Semantic segmentation and candidate generation.}
Given the instruction, branch specification, and textual record of motion, aperture, object poses, and projected points, Qwen divides the demonstration into contiguous subtask intervals $[b_k,e_k)$ covering every action-bearing observation. For each interval, it generates atomic motion, pointing, gripper-trace, and hybrid commands that preserve the relevant branch requirements. The candidate set also includes the original task instruction and the subtask instruction. Offline generation uses thinking disabled and a 12,000-token output limit.

\paragraph{Grounding and validation.}
Qwen selects the entity and frame for each spatial reference; the annotation program inserts the measured coordinates. For example, a command to approach the left edge of the laptop can identify the laptop by its visible-box center. References may use observations through the endpoint of the subtask. A trace contains two to five chronological points from one gripper, and each command contains at most five coordinate pairs. We check interval coverage, reference validity, coordinate bounds, and prompt length, repair or reject invalid annotations, and review a sample for semantic accuracy.

\subsection{Policy Training}

\paragraph{Text conditioning.}
For each training window, we sample a command uniformly from the individual candidates for the subtask containing its first frame. The command replaces the task prompt; the action horizon follows the usual window construction even when the window crosses a subtask boundary. Both policies use the observations, action targets, trainable parameters, and objectives described in Appendix~\ref{sec:policy_training}. For $\pi_{0.5}$, the command and state are tokenized together.

Training and inference share a model-specific coordinate renderer. For $\pi_{0.5}$, the $320\times240$ head image is resized to $224\times168$ and padded vertically by 28 pixels on each side. Thus, head-grid coordinates become $x'=x$ and $y'=\operatorname{round}(3y/4+255/8)$ on the policy's 0--255 canvas grid. For OpenWAM, the head view occupies the upper $256\times320$ region of the $384\times320$ three-view canvas, giving $x'=x$ and $y'=\operatorname{round}(2y/3)$ on its 0--255 grid. Coordinates enter the policy as text tokens; the RGB inputs contain no overlays. Both policies use the augmentation settings in Appendix~\ref{sec:policy_training}.

\begin{table}[t]
    \centering
    \small
    \setlength{\tabcolsep}{3pt}
    \renewcommand{\arraystretch}{1.05}
    \caption{Training settings for Steerable $\pi_{0.5}$. Normalization uses only the training split.}
    \label{tab:steerable_training}
    \begin{tabular}{@{}p{0.35\columnwidth}p{0.60\columnwidth}@{}}
        \toprule
        Setting & Value \\
        \midrule
        Initialization & RoboTwin-pretrained $\pi_{0.5}$ \\
        Training data & 2,520 episodes; task-balanced window sampling \\
        Training budget & 50,000 optimizer updates \\
        Hardware / batch & 8 A100 GPUs; global batch 8 \\
        Seed / workers & 0 / 8 \\
        Observations & Head and two wrist RGB views; 14D joint state \\
        Action representation & Joint deltas and absolute gripper targets; 14D padded to 32D \\
        Horizon / start stride & 50 / 2 \\
        Optimizer & AdamW; $\beta=(0.9,0.95)$; $\epsilon=10^{-8}$ \\
        Weight decay / clip & $10^{-10}$ / gradient norm 1.0 \\
        Learning rate & 1K warmup to $10^{-5}$; cosine decay to $10^{-6}$ at 20K; constant thereafter \\
        EMA / prompt limit & 0.99 / 200 tokens including state \\
        \bottomrule
    \end{tabular}
\end{table}

\paragraph{Training schedule.}
We train both Steerable policies for 50K updates on the clear training split, compared with 30K updates for the task-only policies. Each uses its task-only RoboTwin initialization and training-split normalization. Steerable $\pi_{0.5}$ retains the task-balanced window sampling and optimizer schedule throughout training (Table~\ref{tab:steerable_training}).

\subsection{Online Steering and Evaluation}

\paragraph{High-level controller.}
Every 20 actions, starting at step 0, Qwen receives the original instruction, the current head RGB image, and the episode history of head images and commands. It returns a semantic subtask, atomic motion, pointing command, or hybrid command on the final output line. The controller operates without parameter updates. Gripper-trace commands are used only during training.

\paragraph{Localization and execution.}
For a spatial reference, a second Qwen call locates the target in the current head image. The system maps the result to the head grid and then to the policy canvas. If the locator returns null coordinates, the command retains its verbal descriptions without coordinates. Command generation enables thinking with a 4,096-token output limit; localization disables thinking and returns structured coordinates with a 256-token limit. A normally completed generation with reasoning but no final text is retried at most once.

The policy receives the command, RGB observations, and proprioceptive state. It executes the first 20 predicted actions, or the remaining budget if shorter, before Qwen issues the next command.

\paragraph{Test protocol.}
We evaluate each final 50K checkpoint on all 861 test episodes under each appearance condition using Appendix~\ref{sec:benchmarking_protocol}. In Table~\ref{tab:vla_results} of the main paper, both models achieve lower average joint success with online steering than with the original task instruction under both clear and randomized appearance.

\FloatBarrier

\section{Implementation Details of Inference-only Agents}
\label{sec:agent_implementation}

We evaluate GPT-6 Astra and Qwen3.8-Flash on all 861 \texttt{test\_clear} episodes (Table~\ref{tab:agent_results} of the main paper). Each model selects skills through the same interface and uses the same fixed RGB-D executor, without parameter updates.

\subsection{Agent Interface}

At each decision stage, the model receives the task instruction, current RGB observations, robot state, and action history. It chooses a skill, specifies parameters such as the active arm, and identifies targets in the camera images. A bottle placement, for example, can require separate image points for the cap, the adjacent tabletop, and the destination surface.

\subsection{RGB-D Skill Execution}

\paragraph{Geometric grounding.}
The executor reads the depth at the selected pixel $(u,v)$ in its reference view. Given camera-axis depth $d$ and intrinsic matrix $K$, the target in camera coordinates is
\begin{equation}
    \mathbf p_{\mathrm{cam}} = d K^{-1}[u,v,1]^{\mathsf T},
\end{equation}
Camera extrinsics transform this point into the robot execution frame.

\paragraph{Skill execution and feedback.}
The motion planner uses the resulting targets and skill parameters to execute motion segments and gripper operations. The model then receives updated images and action history and can select another skill or stop. Both agents use the task and instruction checks in Appendix~\ref{sec:task_design_details}; the reported JR and TR weight the eight tasks equally.

\begin{strip}
\section{Additional Policy Evaluations}
\label{sec:zero_shot_ood}

We extend the policy evaluation to LingBot-VA~\cite{lingbot_va}, a video-action world model, and GR00T-N1.7~\cite{groot_n17}, a vision-language-action model. Table~\ref{tab:additional_policy_results} reports zero-shot transfer and post-training results under clear and randomized appearance.

\begingroup
    \centering
    \small
    \setlength{\tabcolsep}{3pt}
    \captionsetup{hypcap=false}
    \captionof{table}{Additional policy evaluation. Cells report JR/TR (\%); averages weight tasks equally. Each evaluation covers 861 episodes. The \texttt{c2c} and \texttt{c2r} columns use the final 30K checkpoints trained on clear demonstrations; GR00T uses global batch 64.}
    \label{tab:additional_policy_results}
    \begin{tabular*}{\textwidth}{@{\extracolsep{\fill}}l*{8}{c}@{}}
        \toprule
        & \multicolumn{4}{c}{\texttt{test\_clear}} & \multicolumn{4}{c}{\texttt{test\_random}} \\
        \cmidrule(lr){2-5}\cmidrule(lr){6-9}
        & \multicolumn{2}{c}{Zero-shot transfer} & \multicolumn{2}{c}{\texttt{c2c}} & \multicolumn{2}{c}{Zero-shot transfer} & \multicolumn{2}{c}{\texttt{c2r}} \\
        \cmidrule(lr){2-3}\cmidrule(lr){4-5}\cmidrule(lr){6-7}\cmidrule(lr){8-9}
        Task & LingBot-VA & GR00T & LingBot-VA & GR00T & LingBot-VA & GR00T & LingBot-VA & GR00T \\
        \midrule
        Place bottle           & 0.0/1.9  & 0.0/0.0 & 99.0/99.0  & 26.9/26.9 & 0.0/0.0  & 0.0/0.0  & 98.1/98.1  & 25.0/25.0 \\
        Shake bottle           & 0.0/42.3 & 0.0/0.0 & 0.0/100.0  & 0.0/29.8  & 0.0/37.5 & 0.0/0.0  & 2.9/98.1   & 0.0/33.7 \\
        Orient fan             & 1.0/1.0  & 0.0/0.0 & 16.3/16.3  & 0.0/0.0   & 1.0/1.0  & 0.0/0.0  & 20.2/20.2  & 0.0/0.0 \\
        Select switch          & 0.0/0.0  & 0.0/0.0 & 3.9/92.2   & 0.0/44.1  & 2.9/6.9  & 0.0/11.8 & 0.0/54.9   & 0.0/22.5 \\
        Transport via divider  & 0.0/0.0  & 0.0/0.0 & 63.2/100.0 & 14.0/38.6 & 0.0/0.0  & 0.0/0.0  & 54.4/100.0 & 41.2/61.4 \\
        Open laptop            & 6.8/8.5  & 0.9/3.4 & 64.1/70.1  & 7.7/8.5   & 6.0/8.5  & 0.9/3.4  & 38.5/49.6  & 1.7/2.6 \\
        Sort blocks in order   & 0.0/0.0  & 0.0/0.0 & 34.3/67.6  & 0.9/1.9   & 0.0/0.0  & 0.0/0.0  & 15.7/73.1  & 0.0/0.0 \\
        Press buttons in order & 0.0/0.0  & 0.0/0.0 & 56.5/77.8  & 0.0/13.9  & 0.0/0.0  & 0.0/0.0  & 15.7/35.2  & 0.0/11.1 \\
        \midrule
        Average                & 1.0/6.7  & 0.1/0.4 & 42.2/77.9  & 6.2/20.5  & 1.2/6.7  & 0.1/1.9  & 30.7/66.1  & 8.5/19.5 \\
        \bottomrule
    \end{tabular*}
\par
\endgroup
\end{strip}

\paragraph{Data and initialization.}
For post-training, both models use the same 2,520 clear demonstrations, original task instructions, and head and wrist RGB views. LingBot-VA starts from \texttt{Robbyant/lingbot-va-posttrain-robotwin}, and GR00T starts from \texttt{nvidia/GR00T-N1.7-3B} through its custom-embodiment interface. Normalization statistics are computed from the RoboInF training split. Zero-shot transfer uses the public checkpoints and their official normalization statistics without RoboInF updates. LingBot-VA retains its RoboTwin interface; GR00T uses its pretrained XDOF embodiment through a heuristic XDOF-to-Aloha adapter.

\paragraph{Post-training.}
Both runs use eight A100-80GB GPUs, BF16, AdamW, and 30K updates. LingBot-VA uses a global batch of eight complete episodes, learning rate $10^{-5}$ with ten warmup steps followed by constant learning rate, and weight decay 0.1. Its transformer learns joint video/action flow matching from pre-encoded latents and text features. GR00T uses a global batch of 64 windows of 40 actions, peak learning rate $10^{-4}$, 1,500 warmup steps, cosine decay, and weight decay $10^{-5}$. It updates projection, attention, and action modules with the vision-language backbone frozen. We evaluate the final 30K checkpoints.

\paragraph{Post-trained policy inference.}
LingBot-VA preserves temporal history and KV cache, generating 16 actions initially and 32 thereafter. Real images arrive every four executed actions while cached actions are consumed. Video/action denoising uses 25/50 steps with guidance scales 5/1. Its adapter outputs absolute end-effector poses and gripper commands. GR00T receives current images and 14D joint-and-gripper state, predicts 40 actions with four denoising steps, and executes 20 before replanning. Relative arm targets and absolute gripper targets are decoded into absolute joint-and-gripper commands.

\newpage
\paragraph{Evaluation protocol.}
We use the same test split and paired appearance conditions as the main paper, evaluating 861 episodes per condition. We report JR/TR with equal weight assigned to each of the eight tasks.

\newpage
\paragraph{Results.}
The post-trained policies achieve higher average scores than their zero-shot transfer configurations under both appearance conditions, but a substantial gap between task and joint success remains. Under clear appearance, LingBot-VA reaches 77.9\% TR and 42.2\% JR, while GR00T reaches 20.5\% TR and 6.2\% JR. This gap persists under randomized appearance. For Sort blocks in order, LingBot-VA achieves JR/TR of 34.3/67.6\% under clear appearance and 15.7/73.1\% under randomized appearance: the order check identifies successful final arrangements that violate the requested sequence. These results reinforce the main finding that task success alone misses violations of execution requirements.

\FloatBarrier

\end{document}